%% file: main.tex
\documentclass{article}

\usepackage{microtype}
\usepackage{graphicx}
\usepackage{subfigure}
\usepackage{booktabs} 

\usepackage{hyperref}

\usepackage[accepted]{icml2024}

\usepackage{amsmath}
\usepackage{amssymb}
\usepackage{mathtools}
\usepackage{amsthm}

\usepackage{algorithm}
\usepackage{algorithmic}

\usepackage[capitalize,noabbrev]{cleveref}

\theoremstyle{plain}

\theoremstyle{definition}

\theoremstyle{remark}

\usepackage[disable,textsize=tiny]{todonotes}

\usepackage{caption}

\icmltitlerunning{Conditionally-Conjugate Gaussian Process Factor Analysis for Spike Count Data via Data Augmentation}

\input{commands}
\begin{document}

\twocolumn[

\icmltitle{Conditionally-Conjugate Gaussian Process Factor Analysis for Spike Count Data via Data Augmentation}



\icmlsetsymbol{equal}{*}

\begin{icmlauthorlist}
\icmlauthor{Yididiya Y. Nadew}{yyy}
\icmlauthor{Xuhui Fan}{xxx}
\icmlauthor{Christopher J. Quinn}{yyy}
\end{icmlauthorlist}

\icmlaffiliation{yyy}{Department of Computer Science, Iowa State University, Ames, IA, USA}
\icmlaffiliation{xxx}{School of Computing, Macquarie University, Sydney, Australia}

\icmlcorrespondingauthor{Christopher J. Quinn}{cjquinn@iastate.edu}

\icmlkeywords{Machine Learning, ICML}

\vskip 0.3in
]



\printAffiliationsAndNotice

\input{sections/abstract}
\input{sections/introduction}

\input{sections/model}

\input{sections/inference}
\input{sections/experiments}
\input{sections/impact-statement}

\bibliography{refs}
\bibliographystyle{icml2024}

\newpage
\appendix
\onecolumn
\input{sections/appendix}

\end{document}

%% file: commands.tex
\usepackage[english]{babel}
\usepackage[utf8]{inputenc}

\usepackage{colortbl,hhline}
\usepackage{soul}
\usepackage{hyphenat}

\usepackage{graphicx}
\usepackage{amsmath}

\usepackage{todonotes}
\usepackage{bibentry}

\newcommand{\bTheta}{\boldsymbol{\Theta}}

\newcommand{\bX}{\boldsymbol{X}}
\newcommand{\bW}{\boldsymbol{W}}
\newcommand{\bK}{\boldsymbol{K}}
\newcommand{\bU}{\boldsymbol{U}}

\newcommand{\br}{\boldsymbol{r}}
\newcommand{\bz}{\boldsymbol{z}}
\newcommand{\bw}{\boldsymbol{w}}

\newcommand{\bZ}{\boldsymbol{Z}}

\newcommand{\btau}{\boldsymbol{\tau}}
\newcommand{\bbeta}{\boldsymbol{\beta}}
\newcommand{\bFsmall}{\boldsymbol{f}}

\newcommand{\bY}{\boldsymbol{Y}}
\newcommand{\bOm}{\boldsymbol{\Omega}}

\newcommand{\bbR}{\mathbb{R}}

\newcommand{\bs}[1]{\boldsymbol#1}

\DeclareMathOperator{\Tr}{Tr}
\DeclareMathOperator{\KL}{KL}

\DeclareMathOperator{\PG}{PG}
\DeclareMathOperator{\diag}{diag}
\DeclareMathOperator{\E}{\mathbb{E}}
\DeclareMathOperator{\const}{const}

\DeclareMathOperator{\PIG}{P-IG}
\DeclareMathOperator{\PTN}{PTN}

\definecolor{orange}{rgb}{1,0.5,0}

%% file: sections/abstract.tex
\begin{abstract}

    Gaussian process factor analysis (GPFA) is a latent variable modeling technique commonly used to identify smooth, low-dimensional latent trajectories underlying high-dimensional neural recordings. Specifically, researchers model spiking rates as Gaussian observations, resulting in tractable inference. Recently, GPFA has been extended to model spike count data.  
    However, due to the non-conjugacy of the likelihood, the inference becomes intractable. Prior works rely on either black-box inference techniques, numerical integration or polynomial approximations of the likelihood to handle intractability.
    To overcome this challenge, we propose a conditionally-conjugate Gaussian process factor analysis (ccGPFA) resulting in both analytically and computationally tractable inference for modeling neural activity from spike count data. 
    In particular, we develop a novel data augmentation based method that renders the model conditionally conjugate. Consequently, our model enjoys the advantage of simple closed-form updates using a variational EM algorithm. 
    Furthermore, due to its conditional conjugacy, we show our model can be readily scaled using sparse Gaussian Processes and accelerated inference via natural gradients. 
    To validate our method, we empirically demonstrate its efficacy through experiments. 

\end{abstract}

%% file: sections/introduction.tex
\section{Introduction} \label{Introduction}

In neuroscience, recent advances in recording techniques have made large-scale neural recordings ubiquitous. 
Common techniques such as neural probes enable simultaneous recording from activity of population of neurons \cite{steinmetz2021neuropixels}. 
Analyzing such high-dimensional data is a challenging statistical problem. 
A recent line of works adopts a generative view using latent variable models (LVMs). 
These methods assume the activity of a neural population lies in a low-dimensional subspace and thus can be captured with a small number of latent variables. 
Such low-dimensional representations can provide insight through  encoding internal neural activity \cite{yu2008gaussian} and also capture relevant information 
 to decode external behaviour such as motor activities \cite{glaser2020machine}.
 
Generally, these models range from classical techniques such as principal component analysis (PCA) and factor analysis (FA) to more advanced methods such as linear dynamical systems (LDS) \citep{semedo2014extracting, gao2015high} and Gaussian process (GP) based models \cite{yu2008gaussian}. 
Unlike classical methods, GP-based models assume the latent variables follow a smooth temporal structure. 
Extending the idea of factor analysis to time series, \citet{yu2008gaussian} proposed a novel method called Gaussian process factor analysis (GPFA). 
GPFA couples dimensionality reduction and temporal smoothness of Gaussian processes in a probabilistic model.  The observations are modeled as conditionally Gaussian, leading to tractable updates within  expectation-maximization (EM) based inference.   

Building on \cite{yu2008gaussian} for Gaussian observations, %
researchers have proposed GPFA variants for count observation models, such as for Poisson distributions or negative binomial distributions, and developed corresponding inference methods, %
to more accurately model spike count data \citep{keeley2020identifying,jensen2021scalable,keeley2020efficient}. 
While models with  discrete distributions of the observations can provide a better fit than Gaussian observation models, they are non-conjugate, which %
makes Bayesian inference intractable. 

To deal with this non-conjugacy, %
Fourier-domain black-box variational inference (BBVI) \citep{keeley2020identifying} and  numerical integration methods \cite{jensen2021scalable} have been proposed, though such approaches have the potential to result in unstable and inaccurate approximations. 
Furthermore, such methods may require complicated settings such as %
learning rate tuning to ensure convergence and annealing techniques to effectively balance model exploration and model selection. 
\citet{keeley2020efficient} leveraged polynomial approximate log-likelihood (PAL) to obtain a marginal likelihood. In particular, this method approximates the non-linear terms in the likelihood using polynomial functions. While this approach yields conjugacy over the latent process, it is not a fully Bayesian scheme since it assumes the factor loading values as parameters and not random variables, making it prone to over fitting. Furthermore, in order to get the polynomial coefficients, the method performs least square solutions over the entire data, posing scalability issues.   
In this work, we present a conditionally-conjugate Gaussian process factor analysis~(ccGPFA) model that can form conjugate models for spike count data. 
It includes an augmenting set of auxiliary variables that render the model conditionally conjugate. 
In contrast to commonly applied P\'{o}lya-gamma augmentation \cite{pillow2012fully,klami2015polya,soulat2021probabilistic}, our method does not require that some variables %
be fixed or learned as model parameters. 
\citet{soulat2021probabilistic} attempted such augmentation for a latent variable model, but relies on moment matching approximations to learn  a parameter  
that controls dispersion of spike counts. 
As demonstrated by authors, despite its effectiveness for over-dispersed data, it poses issues for under-dispersed data. 
 Our approach allows for a conjugate inference for all of the model variables, including the negative binomial dispersion parameter.    
We employ an efficient variational expectation-maximization (EM) algorithm to derive simple closed-form updates for the model.
In short, our contributions are summarized as follows 
\begin{itemize}
    \item We implement a data augmentation technique to make GPFA models conditionally conjugate for spike count data.  

    \item Leveraging the conditional conjugacy, we develop  efficient coordinate ascent inference updates where the posterior of all variables, including the dispersion parameters, are available in closed form.
    
    \item 
    To make inference computationally efficient,  %
    we extend the model and inference method,  incorporating sparse Gaussian process priors and accelerating inference via natural gradients. 
    \item Lastly, we demonstrate the efficiency and efficacy of our model in experiments. %
\end{itemize}

\subsection*{Related works}
The closest related works are \cite{keeley2020identifying,jensen2021scalable,keeley2020efficient}, which we highlighted in the introduction.  We briefly expand on related works here.  See \cref{table:related-works} for a summary of key properties and see \cref{sec:related-extended} %
for a longer discussion.

The standard GPFA model \citet{yu2008gaussian} assumes Gaussian likelihood of the observations.    Due to the nature of spike count data, most subsequent works extend the GPFA model to handle non-conjugate likelihoods such as Poisson and negative binomial.   

\citet{keeley2020identifying} employ techniques such as black-box variational inference (BBVI) to handle non-conjugacy. 
Despite their flexibility, BBVIs do not exploit the structure of model, relying on high variance Monte Carlo estimates. 
In addition they are sensitive to the choice of hyperparameters \cite{locatello2018boosting}. 

\citet{keeley2020efficient} follow an alternative approach using a polynomial approximation of the non-linear terms in the likelihood. 
This transforms the likelihood into a %
quadratic form which makes marginalization of variables easier. 
While this approach yields conjugacy over the latent process, it is not a fully Bayesian scheme since it treats the factor loading values as parameters and not random variables, making it prone to over fitting. Furthermore, in order to get the polynomial coefficients, the method performs least square solutions over the entire data, posing scalability issues.     

\citet{lineartimeGP-ICML23} combined the Hida-Matern Kernels and conjuage computational variational inference to develop latent GP models for neural spikes. However, their method is unable to model the under/over-dispersed spike count data. Specifically, they propose non-conjugate inference for Poisson count models. %

Lastly,
we note that one of the challenges in latent variable inference is selecting the number of latent variables. \citet{jensen2021scalable} %
employ an \textit{automatic relevance determination} (ARD) prior to select the number of latent dimensions in a principled way. 
\citet{gokcen2023uncovering} recently proposed an extension of standard %
GPFA \cite{yu2008gaussian} with ARD for modeling activity from multiple areas. %

\begin{table}[t]
\caption{\textbf{Property comparison with relevant GPFA variants}.
Cnt -- models count data ($\sim$ indicates only Poisson count model);
ARD -- automatic relevance determination to select the number of latents; CF -- closed form updates ($\sim$ indicates only some model parameters have closed form); Scl -- computationally scalable; Trl -- repeated trials ($\sim$ indicates implementations can process multiple trials are not designed for repeated trials; see  \cref{appendix:augmentation:repeated-trials}). 
$\dagger$ inference is specialized for approximating the RBF kernel. 
$\ddagger$ is designed for multi-area.
}
\begin{tabular}{ c|c|c|c|c|c } 
 \toprule
 \textbf{Work} &  \textbf{Cnt} & \textbf{ARD} & \textbf{CF}& \textbf{Scl} &  \textbf{Trl}\\ 
 \toprule
 \citet{yu2008gaussian} &  &   & $\sim$  & & $\sim$\\
 \citet{keeley2020identifying}  & $\sim$ &  &  & & \checkmark \\ %
 \citet{keeley2020efficient}  & \checkmark &   &   & & \checkmark \\
 \citet{jensen2021scalable} $\dagger$  & \checkmark & \checkmark  &  & \checkmark &  $\sim$ \\

   \citet{gokcen2023uncovering} $\ddagger$ &  & \checkmark & \checkmark & & \checkmark \\
 
 Ours   & \checkmark  &  \checkmark & \checkmark & \checkmark & \checkmark \\ 
 \bottomrule
\end{tabular}

\label{table:related-works}
\end{table}

%% file: sections/model.tex
\section{GPFA Model} \label{section:model}

In this section, we formally introduce our conditionally-conjugate Gaussian Process Factor Analysis~(ccGPFA) model for spike count data. 

\subsection{Negative Binomial Modeling for Spike Counts}
We consider the problem of modeling non-negative spike count data. 
Let $\bY \in \mathbb{N}^{N \times T}$ represent the spike counts of $N$ simultaneously recorded neurons over an interval partitioned into $T$ time steps (bins). 
Let $y_{n,t}$ denote the count for neuron $n$ at time step $t$.

We  model
spike counts %
with negative binomial distributions (later we will naturally extend our work to the binomial distribution).  
While Poisson distributions are easier to work with analytically, since the mean and variance are equal they poorly model over (or under) dispersed data.  
Conceptually, the negative binomial distribution models the number of successes in repeated i.i.d. binomial trials before a specified number of failures occur.
For the negative binomial distribution, 
let  $\hat{p}_{n,t}\in[0,1]$ denote the success probability for neuron $n$ at time step (bin) $t$. %
Let $r_n$ model the specified number of failures for neuron $n$.  We refer to $r_n$ as the dispersion parameter since their value can be tuned to account for the ratio of the variance to the mean. Let $\br = \{ r_n \}_{n=1}^N$ denote the set of dispersion parameters for all neurons.  %

Further, %
we model the neural spiking activity as arising from a linear combination of \emph{latent processes} $\bs{f}=\bW\bX+\bbeta\bs{1}^{\top}\in\bbR^{N\times T}$, where $ \bW \in \bbR^{N \times D} (D \ll N)$ is a loading matrix  as the combination coefficients, and $ \bX \in \bbR^{{D} \times T}$ represent $D$-dimensional independent latent processes for $T$ timesteps, and $\bbeta$ is a bias term that represents the base spiking rates of neurons. 
Assuming the success probability $\hat{p}_{n,t}$ is a logistic transformation of $f_{n,t}$, i.e., $\hat{p}_{n,t} = \frac{e^{f_{n,t}}}{1+e^{f_{n,t}}}$ and conditioning on $\bX, \bW, \bbeta$, the neural count data $\bY$ are independently generated across neurons and time and their joint distribution can be factorized as:

\begin{align}
\hspace{-0.2cm}p(\bY| \bW, \bX, \bbeta, \br) = & \prod_{n, t} \text{NegBin}(y_{n,t};r_{n}, \hat{p}_{n,t})
    \label{eq:neg-binomial-gpfa}\\
    = &\prod_{n, t}\frac{\Gamma(y_{n,t} + r_n)[e^{f_{n,t}}]^{y_{n,t}}} {y_{n,t}! \Gamma(r_n)[ 1 +  e^{f_{n,t}}]^{y_{n,t} + r_n }}.
    \label{eq:likelihood-single-count}
\end{align}

In Eq.~\eqref{eq:likelihood-single-count}'s likelihood, it is difficult to find conjugate priors for the following variables since:
\begin{itemize}
    \item the dispersion parameter $r_n$ appears in two Gamma functions. 
    \item the variables $\bX, \bW,$ and $\bbeta$ appear in both the denominator's and the numerator's exponential terms (through $f_{n,t}=[\bW\bX+\bbeta\bs{1}^{\top}]_{n,t}$). 
\end{itemize}

\subsection{ Data Augmentation} \label{subsection:data-augmentation}
In this subsection, we show that by augmenting a set of auxiliary variables, the likelihood in Eq.~\eqref{eq:likelihood-single-count} can be made conditionally conjugate. As a result, we can develop an efficient, fully-Bayesian inference procedure for all variables, including $\{r_n\}_{n=1}^N$.

\textbf{Augmentation for $\{r_{n}\}_{n=1}^N$}
Due to the complex form of the gamma functions, it is hard to find a conjugate prior for $r_n$. 
However, using the following integral representations, identified in \cite{he2019data}, we can transform the gamma function and reciprocal of the gamma function as follows 
\begin{align}
    & \Gamma(y_{n,t} + r_n)  \propto \int_{0}^{\infty} \tau_{n,t}^{(y_{n,t} + r_n)}  e^{-\tau_{n,t}} d\tau_{n,t} \label{eq:aug-r:gamma}  \\
    &\frac{1}{\Gamma(r_n)} =  r_{n} e^{ \gamma r_{n}  } \int_0^\infty e^{ -r_{n}^2 \xi_{n,t}} {\PIG(\xi_{n,t}|0)} d \xi_{n,t} ,  
    \label{eq:augmentation-for-dispersion}
\end{align}
where Eq.~\eqref{eq:aug-r:gamma} represents the marginalization of a gamma variable $\tau_{n,t} \sim \Gamma(y_{n,t} + r_n, 1)$ and  %
Eq.~\eqref{eq:augmentation-for-dispersion} is the convolution of a P\'{o}lya-inverse gamma (P-IG) density. Here, $\gamma \approx 0.577$ denotes Euler's constant. 
These representations are equivalent to augmenting the variables $\tau_{n,t}$ and $\xi_{n,t}$ into the likelihood.
See \cref{sec:background:P-IG} for more details about the P-IG distribution. As will be shown shortly (see \eqref{eq:Yn-like-aug}), this yields conjugacy with respect to $r_n$.

\textbf{Augmentation for $\bs{X}, \bs{W}$ and $\bs{\beta}$ } Inspired by \cite{polson2013bayesian}, we  also augment P\'{o}lya-gamma variables $\{\omega_{n,t}\}_{n=1,t=1}^{N,T}$ into Eq.~\eqref{eq:likelihood-single-count}
and obtain a joint distribution 
\begin{multline}
    p(y_{n,t}, \omega_{n,t}|\bW, \bX, \bbeta, r_n) = \frac{\Gamma(y_{n,t} + r_n)} {y_{n,t}! \Gamma(r_n)}  2^{-(y_{n,t} + r_n)}  \\
    \cdot \exp({ (\frac{y_{n,t} - r_n}{2}) f_{n,t}} - \frac{\omega_{n,t} f_{n,t}^2}{2} ) \PG(\omega_{n,t}| y_{n,t} + r_n, 0),
    \label{eq:likelihood-single-count-augmented:joint}
\end{multline}
where 
$\PG(\omega_{n,t}| y_{n,t} + r_n, 0)$ denotes the P\'{o}lya-gamma distribution \cite{polson2013bayesian} with shape and tilting parameters $y_{n,t}+r_n$ and $0$ respectively.

Therefore upon conditioning on the augmented variables $\{ \omega_{n,t}, \tau_{n,t}, \xi_{n,t} \}$, 
\todo{where do $\tau_{n,t}$, $\xi_{n,t} $ show up?}
dropping factors that are constant with respect to the conditioning variables $\bW$, $\bX$, and $\bbeta$, 
and using notation $z_{n,t} = \frac{y_{n,t} - r_n}{2 \omega_{n,t}}$,  
the likelihood becomes
\begin{multline}    
    p(y_{n,t} | \bW, \bX, \bbeta, \omega_{n,t}, \tau_{n,t}, \xi_{n,t}, r_n )  \\ 
    \propto \mathcal{N}(z_{n,t}|f_{n,t}, \omega_{n,t}^{-1}).
    \label{eq:likelihood-single-count-augmented:cond}  
\end{multline}
\noindent Notice that this likelihood 
is proportional to the probability density function~(pdf) of a corresponding Gaussian variable $z_{n,t}$ with mean $f_{n,t}$ and variance $\omega_{n,t}^{-1}$. This implies, the likelihood is now conditionally conjugate to a Gaussian prior on $\bs{W}, \bs{X}$ and $\bs{\beta}$.  
Generalizing the above derivation for all time steps, and writing $ \bFsmall_{n} = \{ f_{n, t}\}_{t=1}^T, \bOm_n = \diag ( \{ \omega_{n,t} \}_{t=1}^T ), \boldsymbol{z}_n=(\{z_{n,t}\}_{t=1}^T)$, we get a multivariate Gaussian distribution with diagonal covariance (equivalently factorizing into a product of marginal distributions),  
\begin{multline}
    p(\bs{Y}_{n} | \bW, \bX, \bbeta, {\{ \omega_{n,t}, \tau_{n,t}, \xi_{n,t} \}}_{t=1}^T , r_n ) \\
 \propto \mathcal{N}(\boldsymbol{z}_n^\top | \bFsmall_n^\top, \bOm_n^{-1} ). 
\end{multline}

Furthermore, using Eq.~\eqref{eq:augmentation-for-dispersion}, the likelihood $p(\bs{Y}_{n} | \cdot )$ can be simplified as a function of its dispersion variable $r_n$, 
\begin{multline}    
    p(\bs{Y}_{n} | \bW, \bX, \bbeta, {\{ \omega_{n,t}, \tau_{n,t}, \xi_{n,t} \}}_{t=1}^T , r_n )  \\
    \propto r_n^T e^{r_n \sum_t (\log \tau_{n,t}  + \gamma - \log 2   - \frac{1}{2} f_{n,t}) - r_n^2 \sum_t \xi_{n,t} } \label{eq:Yn-like-aug}
\end{multline}

Following \cite{he2019data}, we identify the above expression 
\todo{(minor) did they define Eq.~\eqref{eq:Yn-like-aug} as that? or would Eq.~\eqref{eq:Yn-like-aug} simply be an instance of it; an instance; }
as an un-normalized density of the Power-Truncated-Normal (PTN) distribution. 
The authors show that the gamma distribution can be
a conjugate prior for the above likelihood expression. 
Importantly, we can efficiently estimate the mean of a PTN distribution
which is crucial for our inference method.  See \cref{sec:background:PTN} for more details about this distribution.

\noindent This result will be  important in deriving the closed form updates for our variational distribution. Detailed steps of the augmentation is included in \cref{appendix:augmentation-details}.  %

\subsection{Priors} \label{subsection:priors}
In this subsection, we show that the following choices for prior distributions are indeed (conditionally) conjugate and ease forthcoming inference updates.

\noindent\textbf{Prior for $\bX$}     
We model the prior distribution for the $D$ latent processes,
$p(\bX)$,  as a product of $D$ independent multivariate Gaussian distributions, each with  zero mean and a covariance matrix $\bK_d$ induced by a stationary GP kernel function $k_d(\cdot, \cdot ; \theta_d)$, %
\begin{align}
    p(\bX) = \prod_{d=1}^D \mathcal{N}(\bX_d|0, \bK_d),   [\bK_d]_{t, t'} = k_d(t, t'; \theta_d)
    \label{eq:prior-latents}
\end{align}
where $\theta_d$ denotes kernel specific parameters.  Important example kernels  include the radial basis kernel and the Matérn kernel. See Ch.~4.2 of \cite{rasmussen2006gaussian} for a discussion of different kernels.  %
Any kernel for which parameters can be efficiently updated through automatic differentiation  can be used with our method (see \cref{sec:inference:closed-form}).  
For simplicity, in the following we consider the radial basis kernel, for which  $\theta_d$ is simply a length scale.  

\noindent\textbf{Prior for $\bW$} The prior distribution for the weights $\bW$ are %
modelled as a product of independent multivariate Gaussian distributions along the number of latent dimensions $D$, 
with precisions $\{\tau_d\}_{d=1}^D$ 
(varying among the latent processes %
but shared across neurons),
which in turn are 
modeled with a gamma prior distribution,
\begin{align}
    &p(\bW) = \prod_{n=1}^N \mathcal{N} (\bw_{n} | \boldsymbol{0}, \diag(\frac{1}{\btau})),  \text{and }  p(\btau) = \prod_{d=1}^D \mathcal{G}(a_d, b_d). \label{eq:prior-weights}
\end{align}
This prior over the precision values is a common choice for \textit{automatic relevance determination} (ARD) of latent dimensions (Ch.~6.4 in \cite{bishop1999variational}).
For the latent dimensions where the precision $\btau$ is  large,  the  variance will be small and thus the weights will be concentrated around their (prior) mean of 0, effectively discarding the latent dimension.

We model the prior over the bias terms $\bbeta$ with independent Gaussian distributions.
Similar to placing a prior over the weights' precisions,
we add gamma priors over the common precision parameter $\tau_\beta$ of the distributions,
\begin{align}
    p(\bbeta) = \prod_{n=1}^N \mathcal{N}(\beta_n | 0, \tau_{\beta}^{-1}) \ \text{ and }   \ p(\tau_{\beta}) = \mathcal{G}(c, d), \label{eq:prior-beta}
\end{align}
where $c$ and $d$ are the shape and scale parameters respectively of the gamma distribution.

In addition, as revealed in the augmentation step in Eq.~\eqref{eq:likelihood-single-count-augmented:joint}, we model priors over %
the augmented P\'{o}lya-gamma (PG) variables $\bOm = \{\omega_{n,t}\}$ with 
\begin{align}
    p(\bOm) = \prod_{n=1}^N \prod_{t=1}^T \PG(\omega_{n,t}| y_{n,t} + r_{n}, 0). \label{eq:prior-omega}
\end{align}

\textbf{Prior for $r_n$} 
Following \cite{he2019data}, we use the improper Gamma distribution $\Gamma(1, 0)$ for the prior for $r_n$, i.e $p(r_n) \propto r^{-1}$. This choice yields a proper PTN distribution for its posterior distribution. 

\begin{figure}[t]
    \centering
    \includegraphics[width=0.45\textwidth]{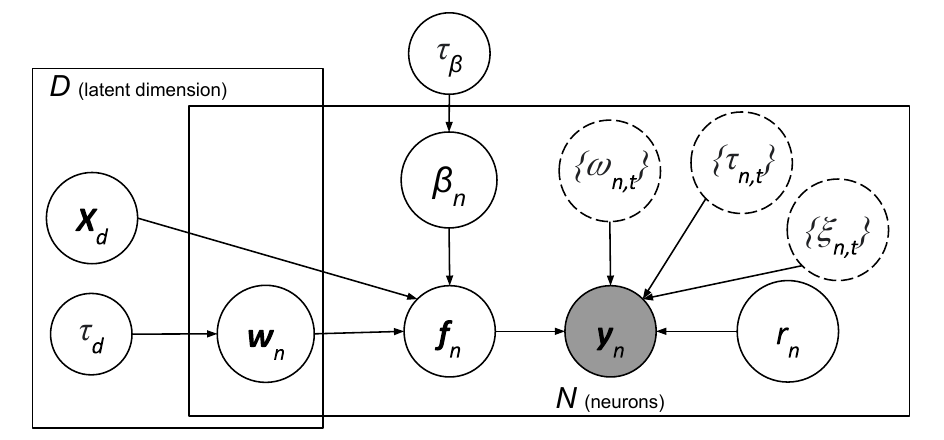}
    \caption{Plate diagram representing our ccGPFA model.  Dashed circles indicate the variable is augmented. 
    }
    \label{fig:plate}
\end{figure}

To this point we have described marginal prior distributions of the latent variables.  After applying the augmentation, 
and the ARD gamma priors, the joint distribution of our model factorizes as   
\begin{multline}
 p(\bs{Y} |\bW, \bX, \bbeta, {\{ \omega_{n,t}, \tau_{n,t}, \xi_{n,t} \}}
 , \{ r_n\}
 )  
 \\ 
 \cdot p(\bW | \btau) p(\bX) p(\bbeta| \tau_{\beta})   p(\btau) p(\tau_{\beta}) \prod_n p(r_n) \\
 \cdot \prod_{n,t} p(\omega_{n,t}) p(\tau_{n,t}) p(\xi_{n,t})
\end{multline}
with additional factorizations arising from equations Eq.~\eqref{eq:prior-weights}, Eq.~\eqref{eq:prior-beta}.
\todo{add the prior distributions of  $\tau_{n,t}$, $\xi_{n,t}$; if defer to appendix is fine, just point reader where to look}
See \cref{fig:plate} for a plate diagram of our model.

\section{Inference}
\label{sec:inference:variational-family} 
In this section, we detail %
our proposed inference procedure for our ccGPFA model. We show the key steps of formula derivations here, whereas more details can be seen in \cref{{appendix:Var-distr}}. %
We use the mean-field variational inference framework to learn the posterior distributions of model variables. Specifically, we approximate the joint augmented posterior $p(\bW, \bX, \bbeta, \btau, \tau_{\beta}, \{\omega_{n,t}, \tau_{n,t}, \xi_{n,t} \}, \{ r_n \} | \bY) \approx q(\bW, \bX, \bbeta, \btau, \tau_{\beta}, \{\omega_{n,t}, \tau_{n,t}, \xi_{n,t} \},  \{ r_n \})$ as a product of variational distributions  
\begin{multline}
 q(\bW) \prod_d q(\bX_d) q(\bbeta)   q(\tau) q(\{ r_n \})\\ 
\cdot q(\tau_{\beta}) q(\{ \omega_{n,t} \}) q(\{ \tau_{n,t} \}) q(\{\xi_{n,t} \})     
\label{eq:inference:variational-family}
\end{multline}

We note that we impose independence over the set of latent processes $\bX = \{ \bX_d \} $ for tractability.

\noindent Beyond the factorization above, we do not impose any parametric distributional assumptions, such as setting $q(\bX_d)$ to be a whitened distribution of the GP priors

%% file: sections/inference.tex
\subsection{Modeling Objective} \label{subsection:scalable-inferece}

Nominally, we would like to optimize the variational distributions
to maximize the marginal log-likelihood of the data $\log p(\bY)$.  
As marginalization of all model variables is intractable, we instead optimize the variational distributions to maximize a lower bound of $\log p(\bY)$. 
Denoting the set of all random variables as $\bTheta = \{\bW, \bX, \bOm, \bbeta, \btau, \tau_{\beta}  {\{ \omega_{n,t}, \tau_{n,t}, \xi_{n,t} \}}, \{ r_n \} \}$, the evidence lower bound (ELBO) $\mathcal{L}$ is 
\begin{align}
    \label{eq:lowerbound}
    \log p(\bY) %
    &\geq \E_{q(\bTheta)} \left[ \log \frac{p(\bY|\bTheta)p(\bTheta)}{q(\bTheta)} \right] =: \mathcal{L}. 
\end{align}
This (standard) bound follows from Jensen's inequality.  For completeness, we show it in \cref{appendix:elbo-derivation}. 

\subsection{Closed form updates } \label{sec:inference:closed-form}

In this work, to optimize the approximate posterior distributions $q(\bTheta)$, 
we employ a variational  Expectation-Maximization (variational EM) algorithm.  See \cref{alg:ccGPFA} for the high-level  pseudocode.  
In the E-step of this algorithm, we apply sequential updates to the distributions in our variational family following the factorization in \eqref{eq:inference:variational-family}.
This step of the algorithm is equivalent to Coordinate Ascent Variational Inference (CAVI). 
We are able to identify closed form updates for this step.  In the M-step,  we maximize the  the marginal lower bound in \eqref{eq:lowerbound} with respect to the model hyperparameters,  such as the characteristic lengthscales $\{\theta_d\}$ of the latent processes $\{\bX_d\}$.

\textbf{Expectation step (CAVI):}
We first show that for our proposed model, we derive closed form coordinate ascent updates that are easy to implement and lead to a computationally efficient procedure.  
We use the notation $\E[q(-\bW)]$ to represent an expectation with respect to to the joint variational distribution of all variables in the variational family except for $\bW$.  
Fixing all variational distributions except one (at a time), we denote  optimal marginal variational distributions with a super-script $*$.  %

By a well known property of mean field variational inference (see Section 10.1 in \cite{bishop2006pattern}), the optimal marginal variational distribution $q^*(\bW)$ (with all others fixed) satisfies $ q^*(\bW) \propto \exp \{ \E_{q(-\bW)} \left [ \log p(\bY, \bTheta) \right] \}$, likewise for other factors in \cref{eq:inference:variational-family}.
We are able to obtain analytic expressions for those optimal marginal variational distributions.   
For the weights $\bW$,
\begin{align*}
    q^*(\bW) &= \prod_{n=1}^N \mathcal{N}(\bw_n | m_n, S_n) \quad \text{with} \nonumber \\
    m_n &= S_n ( \E[\bX] \E[\bOm_n] (\bz_{n}^\top- \E[\beta_n] \boldsymbol{1}) ) \nonumber  \\
    S_n &= ( \diag(\E[\btau]) + \E[\bX \E[\bOm_n] \bX^\top])^{-1}.
\end{align*}
For the base spike rate intensity $\bbeta$,
\begin{align*}
    q^*(\bbeta) &= \prod_{n=1}^N \mathcal{N}(\beta_n | \mu_n, \sigma_n^2) \quad \text{with} \nonumber \\
    \mu_n &= \sigma_n^2 (\bz_n - \E[\bw_{n}] \E[\bX]  ) \E[\bOm_n]  \boldsymbol{1} \nonumber \\ 
    \sigma_n^2 &= 1/{(\E[\tau_{\beta}] + \Tr(\E[\bOm_n]))}.
\end{align*}
\noindent For the gamma precision variables $\{ \btau, \tau_{\beta}\}$,
\begin{align*}
    q^*(\tau_{\beta}) &= \Gamma(c + \frac{N}{2} , d + \frac{1}{2} \sum_{n=1}^N \E[\beta_n^2]) \quad \\
    q^*(\btau) &= \prod_{d=1}^D \Gamma( a_d + \frac{N}{2},  b_d + \frac{1}{2} \sum_{n=1}^N  \E \left [ \bw_{n, d}^2 \right]) 
\end{align*}

For each of the $D$ latent processes $\bX_d$,

\begin{equation} \label{eq:CAVI:Xd}
    \begin{split}
    &q^*(\bX_d) = \mathcal{N}( \hat{\boldsymbol{\mu}}_d, \hat{\bK}_d) \quad \text{with} \\
    &\hat{\boldsymbol{\mu}}_d = \hat{\bK}_d  \bigg (  \E[\bOm_n] \sum_n  \E[\bw_{n,d}] \big( \bz_n^\top  - \E[\beta_n] \boldsymbol{1}
    \\ 
    &\qquad \qquad \qquad  - \sum_{d'\neq d}  \E[\bw_{n, d'}] \E[\bX_{d'}^\top]    \big )\bigg )   \\ 
    & \hat{\bK}_d = (\bK_d^{-1}  + \sum_n \E[w_{n, d}^2] \E[\bOm_n])^{-1} .
    \end{split}
\end{equation}

\noindent For the PG variables $\bOm=\{\omega_{n,t}\}$, %
\begin{align*}
    q^*(\bOm)\! &=\! \prod_{n=1}^N \! \prod_{t=1}^\top \PG(y_{n,t} + r_n, c_{n,t}) 
     \ \text{with} \
    c_{n,t} = \sqrt{\E[f_{n,t}^2]}.
\end{align*}

For the dispersion variables $\{ r_n \}$, 
\begin{multline}    
    q^*(r_n) = \PTN(p, a, b), \text{with }p = T; a = \sum_t \E[\xi_{n,t}];  \\
    b= \sum_t (\E[\log \tau_{n,t}]  + \gamma - \log 2   - \frac{1}{2} \E[f_{n,t}]) \label{eq:variational-updates:dispersion}
\end{multline}

For brevity, we defer details of update rules for the remaining augmented variables to \cref{appendix:Var-distr}.

\textbf{Joint analysis of repeated trials} For simplicity sake, the above derivations show update rules for a single trial. 
However, it can be shown that the above updates can be seamlessly extended to handle repeated trials, such as for vision experiments with repeated trials where the  same visual stimulus.   
See \cref{appendix:augmentation:repeated-trials} for details.

\label{inference:non-identifiability}
\noindent\textbf{ Non-identifiability } In classical GPFA model, there is a problem of 
model non-identifiability due to the interaction of the loading weights and the latent processes \cite{yu2008gaussian}. 
Orthonormalization can be applied on latent states as post processing step. 
In addition to this, in a negative binomial GPFA model, the dispersion variable $r_n$ and the latent function $f_{n,t}$ compete to explain the variance of spike counts as $r_n e^{f_n}(1 + e^{f_n})$ represents the variance of a negative binomial variable. 
To make inference stable in practice, we apply clipping of mean values of the latent function, $\E[f_{n,t}]$,  in the variational update of dispersion variables to a given threshold. 
This is analogous of gradient clipping, a technique common in the machine learning literature.

\textbf{Updating (hyper-)parameters} %
In the M-step, we maximize  the ELBO w.r.t the GP kernel parameters $\{ \theta_d \}_{d=1}^D$. 
In practice, we optimize those parameters with respect to the ELBO \eqref{eq:lowerbound}  using the Adam optimizer algorithm \cite{kingma2017adam}. We do not derive explicit formulas for the gradient here.  In practice, we use the automatic differentiation engine in Pytorch \cite{paszke2019pytorch}. We include the simplification of the ELBO formula in \cref{appendix:elbo}. 
\input{sections/algorithm}

\subsection{Scalable Inference} \label{subsection:scalable-inference}

In \cref{sec:inference:closed-form}, we identified closed form updates for the expectation step.  When the observation length $T$ is large,  evaluating \eqref{eq:CAVI:Xd} can become challenging due to inverting $T \times T$ covariance matrices. 
For inference, we use $M<T$ uniformly-spaced inducing points in modeling the latent GPs $\bX$ to improve efficiency \cite{quinonero2005unifying}.

For each latent dimension $d$, we modify the model by adding set of $M$ inducing points on top of the existing $T$ time points. And for each set of inducing points, we have corresponding $\bU_d$. We first define the joint prior distribution of $\bX_d$ and $\bU_d$
\begin{align}
    p\left (\begin{bmatrix}
      \bX_d  \\
      \bU_d  
    \end{bmatrix} \right ) = \mathcal{N} \left (\begin{bmatrix}
        \boldsymbol{0} \\
        \boldsymbol{0}
    \end{bmatrix} , \begin{bmatrix}
        \bK_{d,tt} & \bK_{d,tm} \\ 
        \bK_{d,mt} & \bK_{d,mm} \\ 
    \end{bmatrix} \right ) , 
\end{align}
where $\bK_{d, tm}$ and $\bK_{d, mt}$ denote cross variances between $\bX_d$ and $\bU_d$.

Using a known property of the multivariate Gaussian distribution, the conditional distribution of $p(\bX_d| \bU_d)$ is given by  
\begin{multline}    
    p(\bX_d| \bU_d) = \mathcal{N} ( \bU_d \bK_{d,mm}^{-1} \bK_{d,mt},  \\
     \bK_{d, tt} - \bK_{d,tm} \bK_{d,mm}^{-1} \bK_{d,mt} ). \label{eq:latent-conditioned-on-inducingpoints}
\end{multline}

\noindent Generalizing for all $D$ latent processes we have 
\begin{align}
     p(\bX,\bU ) = \prod_{d=1}^D  p\left (\begin{bmatrix}
      \bX_d \\
      \bU_d 
    \end{bmatrix} \right ). 
\end{align}
As pointed out in \cite{luttinen2009variational}, assuming the inducing points capture the information in the data well, we can approximate the joint posterior as  
\begin{align}
    q(\bX, \bU) = \prod_{d=1}^D  p(\bX_d| \bU_d ) q(\bU_d). \label{eq:var-dist:inducing-points}
\end{align}

\noindent Careful derivations show all prior derived optimal distributions except for $q^*(\bX_d)$ will remain the same, even after adding the inducing point variables. The updates for the latent processes would be simply marginalizing out $\bU$ from the joint distribution $q(\bX, \bU)$. %
Therefore, it suffices to derive the optimal distributions of the inducing variables. 
\begin{align}
    q^*(\bU_d) \propto \exp \{ \E_{q(-\bU_d)} \left [ \log p(\bY, \bTheta) \right] \} \propto \mathcal{N} (\boldsymbol{m}_d, \boldsymbol{S}_d) , \nonumber 
\end{align}
where 
\begin{align}
   \boldsymbol{S}_d &=  \bigg (\bK_{d, mm}^{-1} + \bK_{d,mm}^{-1} \bK_{d,mt} \bigg ( \sum_n  \E[\bw_{n, d}^2] \E[\bOm_n] \bigg )  \nonumber \\
   &\quad \left.  \bK_{d,mt}^\top \bK_{d,mm}^{-1} \right )^{-1} \nonumber \ \\
   \boldsymbol{m}_d &= \boldsymbol{S}_d \bigg ( \bK_{d,mm}^{-1} \bK_{d,mt}  \E[\bOm_n] \bigg ) \bigg (  \sum_n  \E[\bw_{n,d}] \bigg( \E[\bz_n^\top]    \nonumber  \\
   & -   \sum_{d'\neq d}  \E[\bw_{n, d'}]  \E[\bX_{d'}^\top]  + \E[\bs{\beta_n}] \boldsymbol{1}  \bigg) \bigg).   
\end{align}

With that we can then identify $q^*(\bX_d)$
{ 
\begin{multline}
    q^*(\bX_d) %
     = \mathcal{N}(\bK_{d,tm} \bK_{d,mm}^{-1} \boldsymbol{m}_d,  \\
    \bK_{d,tt} -\bK_{d,tm}\bK_{d,mm}^{-1} (\bK_{d,mm} - \boldsymbol{S}_{d}) \bK_{d,mm}^{-1} \bK_{d,tm}^T).
\end{multline}
}

Using the above update, we effectively avoid an expensive $T \times T$ matrix inversion.

\textbf{Accelerated Inference via Natural Gradients} We note the above updates still require summations through the entire data, which could be prohibitive for analysing long, continuous recordings.
To tackle this challenge, we construct a stochastic version of inference using exponential family distribution properties to derive stochastic natural gradients \cite{hoffman2013stochastic}. 
Unlike other non-conjugate methods, computing the natural gradients does not require computing inverse Fisher information matrix. Deferring details to \cref{appendix:natural-gradients},  we provide the update rules for a natural parameter $\eta$ of a variational distribution,
\begin{align}
    \eta \leftarrow (1 - step\_size ) * \eta + step\_size * \eta_{new }
\end{align}

where the parameter $\eta_{new}$ denotes the noisy natural gradient that uses a mini-batch from the dataset and is appropriately scaled. 
And $step\_size$ is a learning rate hyperparameter which can be close to 1 since we work with natural gradients \cite{hoffman2013stochastic}. %
However, 
Adaptive learning rates \cite{ranganath2013adaptive} can also be applied.

Using sparse GP with $M$ inducing points and mini-batching with batch size $B$, the time complexity of our model is $\mathcal{O}(DM^3 + BM^2)$. 

%% file: sections/algorithm.tex
\begin{algorithm}[t]
    \caption{Inference procedure of ccGPFA}\label{alg:ccGPFA}
    \begin{algorithmic}
        \STATE Input data $\bY$
        \STATE  Initialize latent variables $\bZ \in \bTheta$ 
        \STATE Initialize hyperparameters:  $\{\theta_d\}$ 
        \WHILE{not converged (w.r.t. ELBO \eqref{eq:lowerbound})}
            \STATE 
            \STATE
            \COMMENT{ Expectation step }
            \WHILE{E-step stopping criterion not reached}
                \FOR{ $\bZ \in \bTheta$ }
                    \STATE Update $q^*(\bZ)$ 
                \ENDFOR
            \ENDWHILE
            \STATE
            \STATE
            \COMMENT{ Maximization step }
            \WHILE{M-step stopping criterion not reached}
                \STATE Update length scales $\{\theta_d\}$ 
            \ENDWHILE
        \ENDWHILE
        \STATE Return variational  $q^*(\bs{\Theta})$ and hyperparameters $\{\theta_d\}$
    \end{algorithmic}
\end{algorithm}

%% file: sections/experiments.tex
\section{Experiments}

We compared our method with other baselines on a drifting gratings recording from visual cortex in mice sourced from Allen Brain Observatory data.

\textbf{Dataset} We considered a passive observation segment from the \textit{Allen Brain Observatory: Visual Coding Neuropixels Dataset} \cite{allen-data}. 
Specifically, we analysed a simultaneous recording of 176 neurons from the mouse primary visual (V1) cortex.  The recording was over 75 trials under a drifting gratings stimulus. 
Each trial was 2 seconds long. 
We binned the spike train data into 15 ms bins. 
To evaluate the goodness of fit, we randomly shuffled and split the data into 50 held-in trials and 25 held-out trials.

\textbf{Baseline Methods}
We compared against three common baselines: GPFA \cite{yu2008gaussian}\footnote{https://github.com/NeuralEnsemble/elephant implemented in Python by \cite{elephant18}},
PAL \cite{keeley2020efficient}\footnote{https://github.com/skeeley/Count\_GPFA}, 
and bGPFA \cite{jensen2021scalable}\footnote{https://github.com/tachukao/mgplvm-pytorch}. 
For our method, we considered both binomial and negative binomial observation models.
 For multiple trial data, \cite{keeley2020efficient} and our methods fit a GPFA model with shared set of latents and loading weights across trial.  
 We note that   \cite{yu2008gaussian} and \cite{jensen2021scalable}, however, 
 learn different latent processes for each trial (with a shared covariance matrix; see \cref{appendix:augmentation:repeated-trials} for details).
 In these experiments, to evaluate their fitted models on test data, %
 we computed an average of the latents learned on training data trials.
 We monitored convergence of the training loglikelihood as a common stopping criteria.

\textbf{Metrics} We tested goodness-of-fit of the inferred spike count probabilities and dispersion parameters by computing test log likelihood on held-out trials, computing mean and standard error per neuron and time step.

\textbf{Implementation Notes}
In these experiments, since the trial length was short (133 bins), we did not use inducing points or natural gradients (discussed in \cref{subsection:scalable-inference}) to accelerate inference for our models. The inference procedure is implemented using as a PyTorch(v2.0.1) as a base library.

\begin{table}[t]
    \centering    
    \caption{Performance comparison between our method and baslines in terms of test negative log likelihood (normalized by the total number of bins) and run time (in seconds). }
    \begin{tabular}{c|c|r}
        \toprule
        \textbf{Method } & \textbf{Test NLL} & \textbf{Run time}  \\
        \midrule
         \citet{yu2008gaussian}&   0.3565 $\pm$ 0.0013 & 65.91  \\
         \citet{keeley2020efficient} & 0.3407 $\pm$ 0.0010 & 62.95 \\
         \citet{jensen2021scalable} & 0.3504 $\pm$  0.0013 & 1321.97  \\
         Ours (Binomial) &  0.3390 $\pm$  0.0011 & 9.25 \\
         Ours (NegBinomial)& \textbf{0.3333 $\pm$  0.0010} & \textbf{7.67} \\
         \bottomrule
    \end{tabular}
    \label{tab:experiments:results}
\end{table}

\begin{figure}
     \centering
     \includegraphics[width=0.35\textwidth]{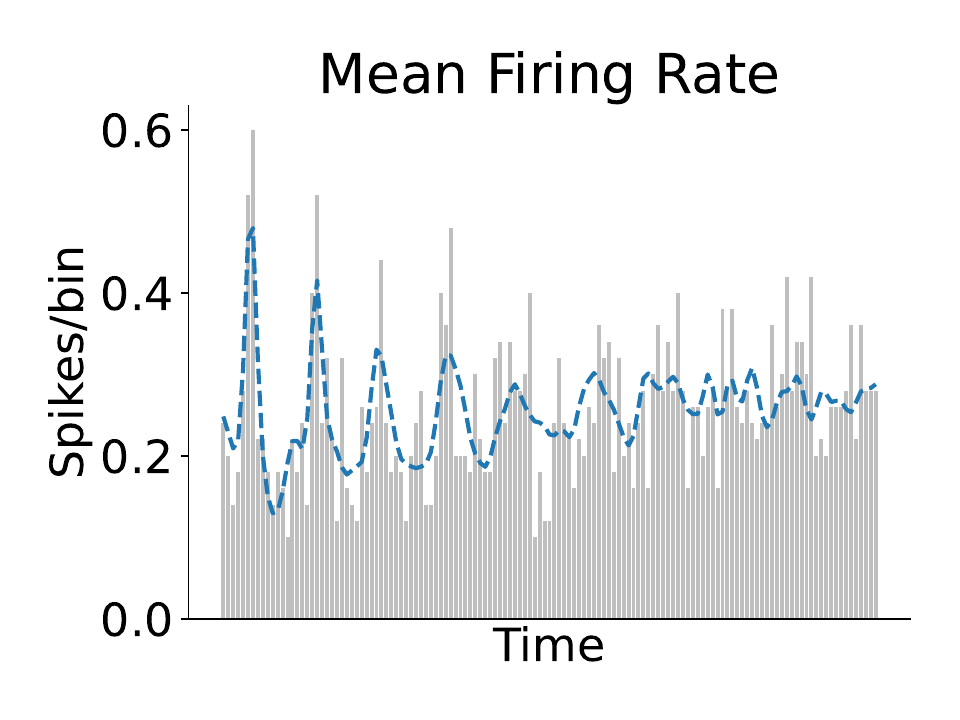}   %
    \includegraphics[width=0.26\textwidth]{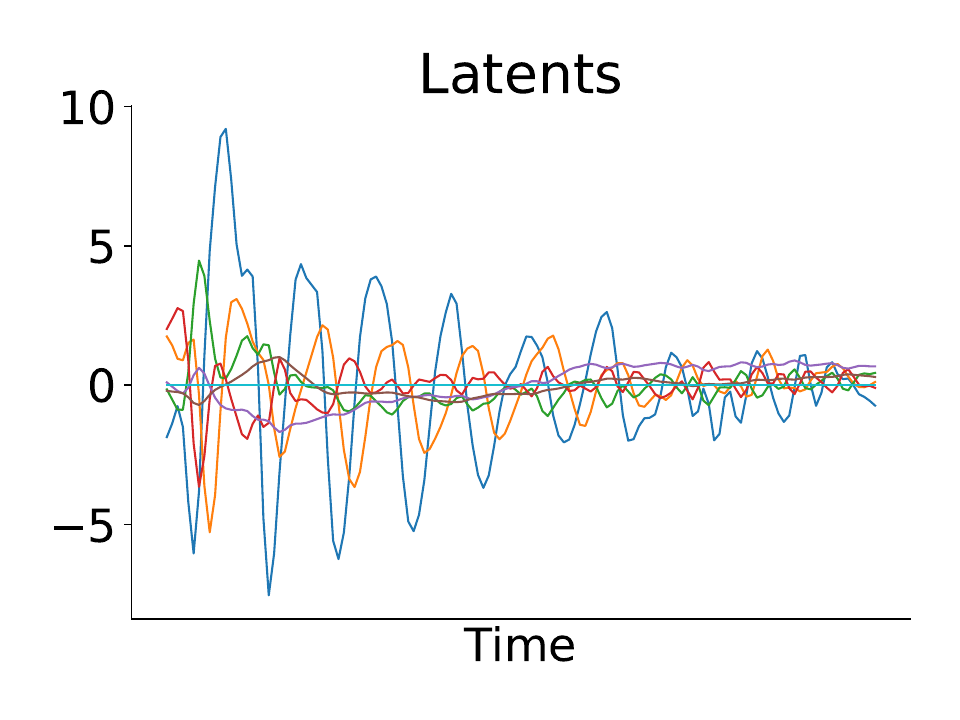}
    \hspace{-.2cm}
    \includegraphics[width=0.22\textwidth]{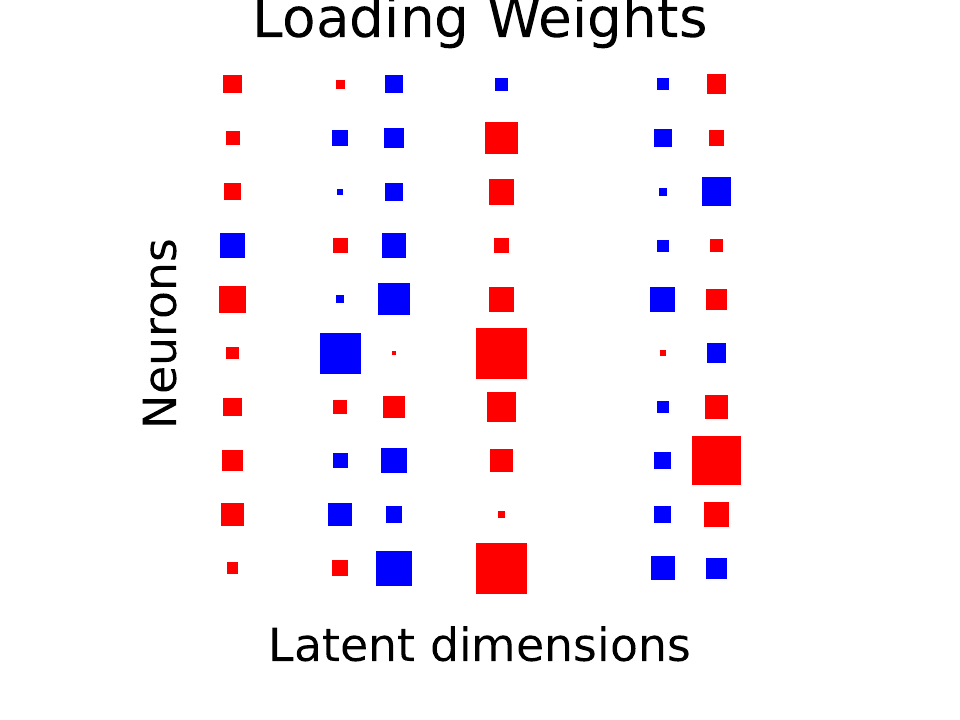}
    \caption{ 
    (\textbf{Top}) inferred mean firing rates by our method of a neuron across time along with peri-stimulus time histogram; 
    (\textbf{Left}) orthonormalized 
    latent processes; 
    \textbf{(Right)} loading weights identified by our method ccGPFA for the first 10 neurons. }
    \label{fig:experiment_results}
\end{figure}

\textbf{Results and Discussion}
We summarize the main results in terms of goodness of fit on test data and run time %
in \cref{tab:experiments:results}.  
Our methods out-performed all of the baselines in terms of both accuracy and speed.  
Both of the models we fit had lower test negative log likelihoods than any baseline, with our negative binomial model performing the best.
Our inference procedure also ran under 10 seconds for both the binomial and negative binomial models, about $1/7$ of the time of the fastest (and also most accurate) baseline, \cite{keeley2020efficient}.

\cref{{fig:experiment_results}} depicts visualizations of our fitted negative binomial model. %
In \cref{{fig:experiment_results}},
the first plot shows a dotted curve representing the mean firing rate  of a neuron, as inferred by our model. The background peri-stimulus histogram represents the neuron's empirical spike rate constructed by computing the mean number of spikes per bin across training trials. 
The  plot of the latent Gaussian processes next to it captures the dominant cyclic pattern also seen in the neural activity.
Furthermore, the Hinton diagram of the weight coefficients (blue means positive, red means negative; size proportional to weight magnitude) shows our model effectively eliminated 4 out of the 10 pre-specified set of latents to yield a concise representation.   

We note \cite{jensen2021scalable} converges slowly and in this experiment no latent processes were fully eliminated, despite employing ARD. 
We speculate that this may in part be due to over parameterization in how \citet{jensen2021scalable} models repeated trials, with distinct latents for each trial that have a shared covariance matrix. An additional factor that may explain why no latents were fully eliminated is that the whitened parameterization \citet{jensen2021scalable} employs constrains the flexibility of the inferred latents, which may necessitate using more latents to model the data than other methods without such constraints. \cite{yu2008gaussian}'s shows relatively poor performance 
in terms of log likelihood, which may partly highlight the importance of count models instead of a Gaussian model.  We also note that \citet{yu2008gaussian}, like \citet{jensen2021scalable}, had unique latents for each trial, so over-parameterization may have been a factor as well. 

%% file: sections/impact-statement.tex
\section{Broader Impact }
This paper presents work whose goal is to advance the field of Machine Learning. There are many potential societal consequences of our work, none which we feel must be specifically highlighted here.

%% file: sections/appendix.tex
We include the following in the supplementary material:
\begin{itemize}
    \item \cref{sec:related-extended}: Related Works - Extended Discussion

    \item \cref{appendix:augmentation-details}:  Augmentation Details

    \item \cref{appendix:Var-distr}: Variational distribution updates

    \item \cref{appendix:elbo}: ELBO Expression
    
    \item \cref{appendix:natural-gradients}: Natural Gradients
    \item \cref{sec:background:important-densities}: Important Densities 

\end{itemize}

\input{sections/appendix/related-works-extended}

\input{sections/appendix/augmentation}

\input{sections/appendix/variational-updates}

\input{sections/appendix/elbo}
\input{sections/appendix/natural-gradients}

\input{sections/appendix/important-densities}

%% file: sections/appendix/related-works-extended.tex
\section{Related Works -- Extended Discussion}
\label{sec:related-extended}

In this section, we extend the review of the related work presented in \cref{Introduction}. 
\subsection{Latent Variable Models}

In general, latent variable models (LVMs) are used to encode neural dynamics from data, which can be later used to decode stimulus inputs and behavioural variables  \cite{glaser2020machine, schimel2021ilqr}.
Broadly, in terms of methodology, LVMs for inferring neural dynamics from data fall into two categories: Gaussian Process based methods \cite{yu2008gaussian, lakshmanan2015extracting,  keeley2020efficient, jensen2021scalable} and auto-regressive methods \cite{yu2005extracting,petreska2011dynamical, semedo2014extracting, pandarinath2018inferring, kim2021inferring}. 
Both lines of works share an underlying assumption that activity of neurons  is driven by a set of shared low-dimensional latent trajectories. 
The main distinction lies on the priors placed for the these factors. 
In addition, the inferred latent state is considered to be discrete-time samples. \citet{duncker2018temporal} extends the GPFA model to continuous time. 
Another common simplifying assumption is that the underlying latent trajectories are assumed independent \textit{a priori}. \citet{rutten2020non} extends the GPFA model to address the limitation. 
In addition, these methods are mostly reserved to observations that are modeled using transformations of linear combinations of latent states. 
However, a recent works \cite{jensen2022beyond} extend GPFA model to non-Euclidean manifold. Recently, in \cite{gokcen2021disentangling, gokcen2023uncovering} extend the standard GPFA to model multiple interacting populations populations. 

In this work, our focus lies on a non-conjugate extension of the GPFA model arising from modeling count data. 
The closest works to ours, \cite{keeley2020efficient, jensen2021scalable}, employ approximate techniques to handle non-conjugacy. 
We present a data augmentation based method to handle non-conjugacy that results in simple closed form solutions. 
\citet{jensen2021scalable} applied variational inference using whitened parameterization tailored to radial basis function (RBF) kernels that resulted in a scalable procedure. 

\subsection{Data augmentation }
 \citet{polson2013bayesian} introduced P\'{o}lya-gamma augmentation to yield model likelihood conditionally conjugate, which is vital in tractable Bayesian inference. 
This technique has been applied to  logistic models \cite{jankowiak2021fast, wenzel2019efficient},  
point process models \cite{zhou2020efficient}, and factor models \cite{pillow2012fully, klami2015polya, soulat2021probabilistic}. 
However, methods using this technique specifically for negative binomial observation model treat dispersion variables as hyperparameters
despite its importance in spike data. 
\citet{he2019data} presents a rich set of augmentation techniques for Gamma based models, including for negative binomial regression model. We extend this augmentation for GP based latent variable model. 
To the best of our knowledge, our work is the first to utilize this augmentation technique on a latent variable model as GPFA. 
In section \cref{inference:non-identifiability}, we elaborate its practical challenges and solution in its application.

\subsection{Sparse Gaussian processes and natural gradients}
Sparse Gaussian processes (GP) have been applied to GP models to mitigate computational complexity, resulting in scalable inference \cite{quinonero2005unifying, titsias2009variational}. 
They are akin to a low-rank approximation of GP covariances, which results in efficient and scalable inference. 
In our model, the latent states are sampled at evenly spaced points in time. 
Therefore, we used fixed inducing points evenly spaced across time. 
When employing Gaussian processes for modeling a set of $n$ data points, using sparse Gaussian processes ($m \ll n$ inducing points) can lead to a reduction in computational complexity from $\mathcal{O}(n^3)$ to $\mathcal{O}(nm^2)$ by inverting an $m \times m$ matrix instead of inverting a larger $n\times n$ matrix.

%% file: sections/appendix/augmentation.tex
\section{Augmentation Details}
\label{appendix:augmentation-details}

Here we show detail of the augmentation steps necessary to make a negative binomial GPFA model conditional conjugate for efficient Bayesian inference.   

\subsection{NegBinomial}
Recall, the likelihood of the observed spike counts $\bs{Y}\in {\mathbb{N}}^{N \times T} $ under a negative binomial GPFA model, given in Eq. \ref{eq:likelihood-single-count}

\begin{align}
    p(\bs{Y} | \cdot) \propto \prod_{n,t} \frac{\Gamma(y_{n,t} + r_n)}{ \Gamma(r_n) } \frac{( e^{f_{n,t}} )^{y_{n,t}} }{ (1 + e^{f_{n,t } } )^{y_{n,t} + r_n } }. \tag{removing $\{y_{n,t}! \}$ as constants}  
\end{align}

To apply Bayesian inference on model variables $\bs{X}, \bs{W}$, $\bs{\beta}$ and $\{r_n\}$, one must either marginalize them out from the joint distribution $p(\bs{Y}, \bs{X}, \bs{W}, \bs{\beta}, \{r_n\} )$ or find prior distributions for which the resulting posterior belongs to a known class of distributions. 
The former entails integration in higher dimensions  which is analytically and computationally intractable. And unfortunately, there is also no known conjugate prior for the given model likelihood.  

Following \cite{he2019data}, we show series of steps to augment the GPFA model to make the likelihood tractable. We first breakdown the likelihood into three parts. 

\begin{align}
    p(\bs{Y} | \cdot)  &\propto \bigg ( \prod_{n,t} \underbrace{\Gamma(y_{n,t} + r_n) }_{\text{term } 1} \bigg ) \bigg ( \prod_{n,t} \underbrace{\frac{1}{\Gamma(r_n) }}_{\text{term } 2} \bigg ) \bigg (\prod_{n,t} \underbrace{\frac{( e^{f_{n,t}} )^{y_{n,t}} }{ (1 + e^{f_{n,t } } )^{y_{n,t} + r_n } } }_{\text{term } 3} \bigg ). 
\end{align}

Commonly used P\'{o}lya-gamma data augmentation techniques \cite{polson2013bayesian} apply an integral identity to term 3 %
into a tractable form yielding Gaussian likelihood over a transformed variable. 
Such augmentations treat the dispersion variable $r_n$ as a parameter, making terms 1 and 2 constant 
upon conditioning.  
Then  $r_n$ is optimized in an outer loop using a common second order optimization techniques \cite{pillow2012fully, soulat2021probabilistic}.  
To apply a fully Bayesian inference on the model, i.e treating $r_n$ as a random variable, we have to also deal with terms 1 and 2. 
And the gamma function does not admit a natural conjugate prior distribution. 
To solve this, \citet{he2019data} identifies the following integrals.

First, we can express the product of gamma functions in term 1 as follows,
\begin{align}
    \prod_{n,t} \Gamma(y_{n,t} + r_n) \propto \prod_{n,t} \int_0^\infty \tau_{n,t}^{(y_{n,t} + r_n) - 1} e^{-\tau_{n,t}} d\tau_{n,t} . \label{eq:appendix:augmentation:tau}
\end{align}

Next, we apply another integral equivalence to the reciprocal gamma function in term 2, 
\begin{align}
    \prod_{n,t} \frac{1}{\Gamma(r_n)} 
    &\propto \prod_{n,t} \int_0^\infty r_{n} e^{ -r_{n}^2 \xi_{n,t} + \gamma r_{n}  } \PIG(\xi_{n,t}|0) d \xi_{n,t} \\
    &\propto  \prod_{n,t} r_{n} e^{ \gamma r_{n}  } \int_0^\infty e^{ -r_{n}^2 \xi_{n,t}} \PIG(\xi_{n,t}|0) d \xi_{n,t} ,     \label{eq:appendix:augmentation:xi}
\end{align}
where \eqref{eq:appendix:augmentation:xi} follows from pulling out constants and  $\PIG(0)$ denotes the PDF of a Polya-Inverse Gamma distribution, 
 a new class of distributions developed in \cite{he2019data}, with tilting parameter 0. 
 In the equation above, $\gamma \approx 0.577$ refers to Euler's constant.

Finally, we apply an integral identity to transform term 3, %
\begin{align}
    \prod_{n,t} \frac{( e^{f_{n,t}} )^{y_{n,t}} }{ (1 + e^{f_{n,t } } )^{y_{n,t} + r_n } }  = \prod_{n,t} 2^{-b} \exp((a- b/2) f_{n,t} ) \int_0^\infty \exp \big \{ - \frac{\omega_{n,t} f_{n,t}^2}{2} \big \} \PG(\omega_{n,t}| b, 0) d \omega_{n,t} ,
    \label{eq:appendix:augmentation:polya-gamma}
\end{align}

where $a=y_{n,t}$ and $b=y_{n,t} + r_n$. Simplifying the expression by removing terms with only $y_{n,t}$ dependence and setting $\kappa_{n,t} = \frac{y_{n,t} - r_n }{2}$,

\begin{align}
    \prod_{n,t} \frac{( e^{f_{n,t}} )^{y_{n,t}} }{ (1 + e^{f_{n,t } } )^{y_{n,t} + r_n } }  \propto \prod_{n,t}  \exp( - r_n \log 2 + \kappa_{n,t} f_{n,t} ) \int_0^\infty \exp \big \{ - \frac{\omega_{n,t} f_{n,t}^2}{2} \big \} \PG(\omega_{n,t}|b, 0) d \omega_{n,t} .
\end{align}

Treating the newly introduced variables as latent auxiliary variables in the model, we identify the joint conditional distribution $p(\bs{Y}, \{ \tau_{n,t}, \xi_{n,t}, \omega_{n,t} \}_{n=1...N,t=1...T} |  \bs{X}, \bs{W}, \bs{\beta}, \{r_n\} )$. For simplicity,  we omit the conditioning variables. 

Gathering the above results, and augmenting the variables into the model, 
\begin{align}
    p(\bs{Y}, \{  &\tau_{n,t}, \xi_{n,t}, \omega_{n,t} \}_{n=1...N,t=1...T} | \cdot ) \propto \bigg (\prod_{n,t}   \tau_{n,t}^{(y_{n,t} + r_n) - 1} e^{-\tau_{n,t}}  \bigg ) \times \bigg ( \prod_{n,t}   r_{n} e^{ -r_{n}^2 \xi_{n,t} + \gamma r_{n}  } \PIG(\xi_{n,t}|0) 
 \bigg )  \nonumber \\
 & \times  \prod_{n,t}  \exp( - r_n \log 2 + \kappa_{n,t} f_{n,t} )  \exp \big \{ - \frac{\omega_{n,t} f_{n,t}^2}{2} \big \} \PG(\omega_{n,t}) .
\end{align}

Conditioning on the augmented variables, we can note the augmented likelihood becomes \textit{conditionally conjugate}. 
By placing Gaussian priors on $\bs{X}, \bs{W}$, and $\bs{\beta}$, their conditional posterior is a Gaussian density up to a constant factor. Similarly, placing a gamma prior on $r_n$, $p(r_n) \sim \Gamma(a_0, b_0)$ yields an exponential conditional posterior. Similar to \cite{he2019data}, we use an improper gamma prior $\Gamma(1, 0)$. 

As a function of $f_{n,t}$, 
\begin{align}
    p_{f_{n,t}}(\bs{Y}  | \{  \tau_{n,t}, \xi_{n,t} \omega_{n,t} \}_{n=1...N,t=1...T}, \cdot ) 
    &\propto  \prod_{n,t}  \exp (\kappa_{n,t} f_{n,t}   - \frac{\omega_{n,t} f_{n,t}^2}{2} )  \nonumber \\
    &\propto \prod_{n,t} \exp( - \frac{1}{2} \omega_{n,t} (\frac{\kappa_{n,t}}{ \omega_{n,t}} - f_{n,t})^2) \nonumber \\
    &\propto \mathcal{N}(z_{n,t} | f_{n,t}, \omega_{n,t}^{-1} ) \tag{ $z_{n,t} = \frac{\kappa_{n,t}}{\omega_{n,t}}$}.
\end{align}

And as a function of ${r_n}$
\begin{align}
    p_{r_n}(\bs{Y}| \{  \tau_{n,t}, \xi_{n,t} \omega_{n,t} \}_{n=1...N,t=1...T} ) \propto \exp \{ \sum_t {r_n}\log\tau_{n,t} + \log r_n - r_n^2 \xi_{n,t} + \gamma r_n -r_n \log2  -\frac{1}{2} r_n f_{n,t} \}. \nonumber 
\end{align}

This result lays the foundation for closed form variational updates in \cref{appendix:Var-distr}. Unlike the original negative binomial likelihood, this augmented likelihood ensures conjugacy to all our priors and is fundamental in deriving closed form updates for all our variables, including the newly augmented ones.

\subsection{Binomial GPFA} As mentioned in the main paper, similar  augmentation can be applied to a GPFA model with Binomial observations. 
Consider the following binomial model, where $k_n$ represents the total number of Bernoulli trials and $p$ is the probability of success, linked to the latent function $f_{n,t}$ via log-odds.    %
\begin{align}
    p(y_{n,t} | \cdot ) &\propto p^{y_{n,t}} (1-p)^{k_n - y_{n,t}} \nonumber \\
    & \propto \frac{(e^{f_n,t})^{y_{n,t}} }{ ( 1 + e^{f_n,t} )^{k_n}}.
\end{align}

By setting $a= y_{n,t}$ and $b=k_n$ and applying the  integral identity in \cref{eq:appendix:augmentation:polya-gamma}, we can restore conjugacy to the model. 
In addition, the common choice of the total number of trials $k_n$ is the maximum number of spikes for neuron $n$ over the length of the recording. Therefore, the value would be fixed and removes the necessity for further augmentation, unlike the negative binomial GPFA model.  

\subsection{Repeated trials}
\label{appendix:augmentation:repeated-trials}

In the following, we extend the problem setup from analysing single trial data to repeated trial data. Let  $\{ \bs{Y}^{m}\}_{m=1...M}$ denote the spike counts from $M$ repeated trials. 
Extending Eq. \ref{eq:likelihood-single-count} to this setup,  our likelihood becomes 

\begin{align}
    p(\{ \bs{Y} \}_m | \cdot)  &\propto \bigg ( \prod_{m,n,t} \underbrace{\Gamma(y_{m,n,t} + r_n) }_{1} \bigg ) \bigg ( \prod_{m,n,t} \underbrace{\frac{1}{\Gamma(r_n) }}_{2} \bigg ) \bigg (\prod_{n,t} \underbrace{\frac{( e^{f_{n,t}} )^{\sum_m y_{m,n,t}} }{ (1 + e^{f_{n,t } } )^{\sum_m y_{m, n,t} + M r_n } } }_{3} \bigg ) 
\end{align}

 Note that we model the observation across trials as arising from shared set of latents and loading weights. 

By setting $a = \sum_m y_{m,n,t}$ and $b= \sum_m y_{m,n,t} + M r_n$, we can apply the integral identity in Eq. \ref{eq:appendix:augmentation:polya-gamma}. 
This is equivalent to augmenting $N \times T$ variables. 
We further augment the model for each product term in $1$ and $2$ following Eqs. \ref{eq:appendix:augmentation:tau} and \ref{eq:appendix:augmentation:xi}. 
After augmentation, the subsequent derivations simply follow. 

Prior works handle repeated trials differently. \cite{yu2008gaussian,jensen2021scalable} learned shared parameters such as GP lengthscales and weights and allowing for varying latents. \cite{keeley2020efficient} , similar to ours, model shared set of latents and weights to model the common structure across trials, which is beneficial in analyzing repeated trials as shown in the experimental results.

%% file: sections/appendix/variational-updates.tex
\section{Variational distribution updates}
\label{appendix:Var-distr}

Using the mean field assumption provided in Eq. \eqref{eq:inference:variational-family}, we derive the optimal distributions, denoted with $*$ superscript, for the variables using coordinate ascent variational inference \cite{bishop2006pattern}. We represent the set of all variables including the augmented variables with $\bs{\Theta}$.

\subsection{Augmented variables }
For augmented gamma variables, $\tau_{n,t}$
\begin{align}
    q^*(\tau_{n,t}) &  \propto \exp \{ \E_{q(-\tau_{n,t} )} [\log p(\bs{Y}, \bs{\Theta})] \} \nonumber \\
    & \propto  \exp \{ \E[ ((y_{n,t} + r_n) - 1) \log  \tau_{n,t}  - \tau_{n,t}] \} \tag{removing constant terms} \nonumber \\
    & \propto \exp \{ ((y_{n,t} + \E[r_n]) - 1) \log  \tau_{n,t}  - \tau_{n,t} \}. \tag{expectation;} 
\end{align}

Recognizing the above as gamma distribution with natural parameters $[ (y_{n,t} + \E[r_n]) - 1, -1]$, the optimal distribution is given by 
\begin{align}
    q^*(\tau_{n,t}) = \Gamma(y_{n,t} + \E[r_n], 1).
\end{align}

For augmented $\PIG$ variables, $\xi_{n,t}$
\begin{align}
    q^*(\xi_{n,t}) &\propto \exp \{ \E_{q(-\xi_{n,t} )} [\log p(Y, \bs{\Theta})] \} \nonumber \\
    & \propto p(\xi_{n,t}; 1) \exp \{ { - \E[r_{n}^2] \xi_{n,t} } \}. \tag{removing constant terms} 
\end{align}

Using the exponential tilting property \cite{he2019data}, we can recognize this as a general class $\PIG$ distribution, 
\begin{align}
    q^*(\xi_{n,t}) = \PIG(\sqrt{\E[r_n^2]}).
\end{align}

For the set of augmented PG variables $\omega_{n,t}$ we can derive the optimal variational distributions as follows, 
\begin{align}
    q^*(\omega_{n,t}) &\propto \exp \{ \E_{q(-\omega_{n,t} )} [\log p(Y, \bs{\Theta})] \} \nonumber \\
    & \propto  \exp \bigg \{ \E \bigg [ \log \frac{( e^{f_{n,t}} )^{y_{n,t}} }{ (1 + e^{f_{n,t } } )^{y_{n,t} + r_n } } \bigg ] \bigg  \}  \tag{removing terms with no $\omega_{n,t}$ dependence } \\
    & \propto  \exp \bigg \{ \E \bigg [ {y_{n,t}}  \log ( e^{f_{n,t}} ) - (y_{n,t} + r_n ) \log  (1 + e^{f_{n,t }  ) } \bigg ] \bigg  \}  \tag{removing terms with no $\omega_{n,t}$ dependence } \\
    & \propto  \exp \bigg \{ \E \bigg [ {y_{n,t}}  \log ( e^{f_{n,t}} ) - (y_{n,t} + \E[r_n] ) \log  (1 + e^{f_{n,t }  ) } \bigg ] \bigg  \}  \tag{applying expectation  w.r.t $r_n$ } \\
    & \propto  \exp \bigg \{ \E \bigg [ \log \frac{( e^{f_{n,t}} )^{y_{n,t}} }{ (1 + e^{f_{n,t } } )^{y_{n,t} + \E[r_n] } } \bigg ] \bigg  \}  \tag{exponent in log; quotient rule } \\
     & \propto  \exp \bigg \{ \E \bigg [ \log \exp \{ - \frac{\omega_{n,t} f_{n,t}^2}{2}  \} \PG(\omega_{n,t}| y_{n,t} + \E[r_n], 0 ) \bigg ] \bigg  \}  \tag{applying the P\'{o}lya-gamma integral identity; dropping constants } \\
    & \propto \PG(\omega_{n,t}| y_{n,t} + \E[r_n], 0) \exp \bigg \{ \E[  - \frac{\omega_{n,t} f_{n,t}^2}{2} ] \bigg  \} \nonumber \tag{simplifying expression }\\
    & \propto \PG(\omega_{n,t}|y_{n,t} + \E[r_n], \sqrt{\E[f_{n,t}^2]}). \tag{exponential tilting}
\end{align}
Note all augmented variables $\{ \omega_{n,t} \}$ do not have interdependence in the updates, hence the updates can be done in parallel fashion. 
The same  holds for $\{ \tau_{n,t}, \xi_{n,t} \}$.

For the repeated trial setting, \ref{appendix:augmentation:repeated-trials}, the updates would  simply differ in the shape parameter $PG(\omega_{n,t}| \sum_{m} y_{m, n,t} + E[r_n], \sqrt{E[f_{n,t}^2]} )$

\subsection{Dispersion variables}

\begin{align}
    q^*(r_n) &\propto \exp \{ \E_{q(-r_n)} [ \log p(\bs{Y}, \bs{\Theta} ) ]\} \nonumber \\
    & \propto \exp \{ \E[ \log p(\bs{Y}|\cdot) p(r_n) ]\} \nonumber \\
    & \propto p(r_n) \exp \{ \E[ \log p(\bs{Y}|\cdot) ]\} \nonumber \\
    & \propto p(r_n) \exp \{ \E[ \log \prod_t \tau_{n,t}^{(y_{n,t} + r_n) -1}  r_n \exp (-r_n^2 \xi_{n,t} + \gamma r_n -r_n \log2 + \kappa_{n,t} f_{n,t}) ]\} \tag{expanding expression by droping terms with no $r_n$ dependence} \nonumber \\    
    & \propto p(r_n) \exp \{ \E[ \sum_t {r_n}\log\tau_{n,t} + \log r_n - r_n^2 \xi_{n,t} + \gamma r_n -r_n \log2  -\frac{1}{2} r_n f_{n,t} ]\} \nonumber \tag{product to summation; $\kappa_{n,t} = \frac{y_{n,t} - r_n}{2}$} \\
    & \propto p(r_n) \exp \{  \sum_t {r_n} \E[\log\tau_{n,t}]+ \log r_n -r_n^2 \E[\xi_{n,t}] + \gamma r_n - r_n \log2  -\frac{1}{2} r_n \E[f_{n,t} ]\} \nonumber \tag{applying expectation}\\
    & \propto p(r_n) r_n^T\exp \{  \sum_t {r_n} \E[\log\tau_{n,t}]-r_n^2 \E[\xi_{n,t}] + \gamma r_n - r_n \log2  -\frac{1}{2} r_n \E[f_{n,t} ]\} \nonumber \\
    & \propto r_n^{T-1}\exp \{ -r_n^2 \bigg ( \sum_t \E[\xi_{n,t}] \bigg ) +  {r_n} \sum_t  \bigg (\E[\log\tau_{n,t}] + \gamma - \log2  -\frac{1}{2} \E[f_{n,t}] \bigg ) \} \nonumber \tag{$p(r_n) \propto 1/r $; simplification}\\
\end{align}

We can recognize the above expression as a Power Truncated Normal ($\PTN$) distribution \cite{he2019data} with parameters 

\begin{align}
    p &= T ,  \\
    a &= \sum_t \E [\xi_{n,t}] \And b = \sum_t \bigg ( \E[\log\tau_{n,t}] + \gamma - \log2  -\frac{1}{2} \E[f_{n,t}] \bigg )  .
\end{align}

The expectation in $E[\xi_{n,t}]$ is given closed form of the moment of a $\PIG$ distribution with tilting parameter $c$ (Theorem 2 in \cite{he2019data} for details)

\begin{align}
    \E[\xi_{n,t}] = \frac{1}{2c} \psi(c + 1) - \psi(1) .
\end{align}

Also by logarithmic expectation of gamma distribution, we can simplify, 
\begin{align}
    \E[\log \tau_{n,t}] = \psi(\alpha) - \log \beta .
\end{align}
where $\alpha, \beta$ are shape and rate parameters of the variational distribution and $\psi$ denotes the digamma function. 

\textbf{Repeated trials setting} For $M$ repeated trials,  

\begin{align}
    p &= M T, \\
     a &= \sum_{m,n,t} \E [\xi_{m, n,t}] \And b = \sum_{m,n,t} \bigg (  \E[\log\tau_{m,n,t}] + \gamma - \log2  -\frac{1}{2} \E[f_{n,t}] \bigg ) 
\end{align}

where $\{ \xi_{m,n,t}, \tau_{m,n,t}\}$ are augmented variables. 

\subsection{Latent processes}

\begin{align}
    q^*(\bs{X}_d) &\propto p(\bs{X}_d) \exp \{ \E_{q(-\bs{X}_d)} [\log p(\bs{Y} | \Theta)]  \} \nonumber \\
    &\propto p(\bs{X}_d) \exp \{ \E \bigg [ -\frac{1}{2}\sum_n (\bs{z}_n - \bs{f}_n) \bs{\Omega}_{n}  (\bs{z}_n - \bs{f}_n)^\top \bigg ]  \} \nonumber \\
    &\propto p(\bs{X}_d) \exp \{ \E \bigg [  -\frac{1}{2}\sum_n -2  \bs{f}_n \bs{\Omega}_{n} \bs{z}_n^\top +   \bs{f}_n \bs{\Omega}_{n} \bs{f}_n^\top \bigg ]  \} \nonumber \tag{distribute and remove constants w.r.t $\bs{X}_d$}\\
    &\propto p(\bs{X}_d) \exp \{ \E \bigg [  -\frac{1}{2} \sum_n    
 \bs{w}_n \bs{X}  \bs{\Omega}_{n} \bs{X}^\top 
 \bs{w}_n^\top - 2 \bs{w}_n \bs{X}  \bs{\Omega}_{n} (  \bs{z}_n^\top   - \bs{\beta}_n \bs{1})  \bigg ]  \} \nonumber \tag{removing constants w.r.t $\bs{X}_d$}\\
    &\propto p(\bs{X}_d) \exp \{ \E \bigg [  -\frac{1}{2} \sum_n \bigg ( \sum_{d'}   
 \bs{w}_{n,d} \bs{X}_d  \bs{\Omega}_{n} \bs{X}_{d'}^\top 
 \bs{w}_{n,d'} \bigg  ) - 2 \bs{w}_{n,d} \bs{X}_d  \bs{\Omega}_{n} (  \bs{z}_n^\top   - \bs{\beta}_n \bs{1})  \bigg ]  \} \nonumber \tag{decomposing summations; removing constants }\\
    &\propto p(\bs{X}_d) \exp \{ \E \bigg [  -\frac{1}{2} \sum_n 
  \bs{X}_d  (\bs{\Omega}_{n} \bs{w}_{n,d}^2) \bs{X}_{d}^\top  - 2 \bs{X}_d  \bs{w}_{n,d} \bs{\Omega}_{n} (  \bs{z}_n^\top   - \bs{\beta}_n \bs{1} -   \sum_{d'}  \bs{X}_{d'} \bs{w}_{n, d'})  \bigg ]  \} \nonumber \tag{rearranging  }\\
    &\propto p(\bs{X}_d) \exp \{  -\frac{1}{2} \bigg ( 
  \bs{X}_d  ( \sum_n  E[\bs{\Omega}_{n}] \E[\bs{w}_{n,d}^2] ) \bs{X}_{d}^\top \nonumber \\  
  &\quad \quad- 2 \bs{X}_d \bigg (  \sum_n \E[\bs{w}_{n,d}] \E[\bs{\Omega}_{n}] (  \bs{z}_n^\top - \E[\bs{\beta}_n] \bs{1} -   \sum_{d' \neq d }  \E[\bs{X}_{d'}] \E[\bs{w}_{n, d'})] \bigg )    \bigg ) \} \nonumber \tag{applying summations along $n$; applying expectations   }\\
    &\propto\exp \{  -\frac{1}{2} \bigg ( 
  \bs{X}_d  \underbrace{(\bs{K}_d^{-1}  + \sum_n  \E[\bs{\Omega}_{n}] \E[\bs{w}_{n,d}^2] )}_{:= \bs{\Phi}_d} \bs{X}_{d}^\top \nonumber \\  
  &\quad \quad- 2 \bs{X}_d \bigg (  \sum_n \E[\bs{w}_{n,d}] \E[\bs{\Omega}_{n}] (  \bs{z}_n^\top - \E[\bs{\beta}_n] \bs{1} -   \sum_{d' \neq d }  \E[\bs{X}_{d'}] \E[\bs{w}_{n, d'})] \bigg )    \bigg ) \} \nonumber \tag{definition of $p(\bs{X}_d)$ \ref{eq:prior-latents} }\\
    &\propto \exp \bigg \{   -\frac{1}{2}    
 \bigg ( \bs{X}_d - \bigg ( \bs{\Phi}_d^{-1}\bigg (  \sum_n \E[\bs{w}_{n,d}] \E[\bs{\Omega}_{n}] (  \bs{z}_n^\top - \E[\bs{\beta}_n] \bs{1} -   \sum_{d' \neq d }  \E[\bs{X}_{d'}] \E[\bs{w}_{n, d'})] \bigg )  \bigg )^\top   \bigg ) \bs{\Phi}_d \nonumber \\
 & \qquad \qquad \bigg ( \bs{X}_d - \bigg ( \bs{\Phi}_d^{-1}\bigg (  \sum_n \E[\bs{w}_{n,d}] \E[\bs{\Omega}_{n}] (  \bs{z}_n^\top - \E[\bs{\beta}_n] \bs{1} -   \sum_{d' \neq d }  \E[\bs{X}_{d'}] \E[\bs{w}_{n, d'})] \bigg )  \bigg )^\top   \bigg )^\top \bigg   \}  \tag{removing constant;  applying expectations; completing the square; }
\end{align}

We can recognize the above expression as a multivariate Gaussian distribution over a random vector $\bs{X}_d$ with mean and variance $\bs{m}_{d}$ and variance $\bs{V}_d$ 

\begin{align}
    q^*(\bs{X}_d) = \mathcal{N}(\bs{m}_d, \bs{V}_d) \nonumber \\
    \bs{m}_d = \bs{V}_d \bigg ( \sum_n \E[\bs{w}_{n,d}] \E[\bs{\Omega}_{n}] (  \bs{z}_n^\top - \E[\bs{\beta}_n] \bs{1} -   \sum_{d' \neq d }  \E[\bs{X}_{d'}] \E[\bs{w}_{n, d'})] \bigg )   \\
    \bs{V}_d = \bs{\Phi}_n^{-1}
\end{align}

\subsection{Base intensities}
\begin{align}
    q^*(\bs{\beta}) &\propto p(\bs{\beta} | \tau_{\beta}) \exp \{ \E_{q(-\bs{\beta}} [\log p(\bs{Y} | \bs{\Theta})]  \} \nonumber \\
    &\propto p(\bs{\beta} | \tau_{\beta}) \exp \{ \E \bigg [ -\frac{1}{2}\sum_n (\bs{z}_n - \bs{f}_n) \bs{\Omega}_{n}  (\bs{z}_n - \bs{f}_n)^\top \bigg ]  \} \nonumber \\
    &\propto p(\bs{\beta} | \tau_{\beta}) \exp \{ \E \bigg [  -\frac{1}{2}\sum_n -2  \bs{f}_n \bs{\Omega}_{n} \bs{z}_n^\top +   \bs{f}_n \bs{\Omega}_{n} \bs{f}_n^\top \bigg ]  \} \nonumber \tag{distribute and remove constants w.r.t $\bs{\beta}_n$}\\
    &\propto p(\bs{\beta} | \tau_{\beta}) \exp \{ \E \bigg [  -\frac{1}{2} \sum_n    
 \bs{\beta}_n \bs{1}^\top  \bs{\Omega}_{n} \bs{1}
 \bs{\beta}_n - 2 \bs{\beta}_n \bs{1}^\top  \bs{\Omega}_{n} (  \bs{z}_n^\top   - \bs{X}^T \bs{w}_n^T)  \bigg ]  \} \nonumber \tag{removing constants w.r.t $\bs{\beta}_n$}\\
    &\propto p(\bs{\beta} | \tau_{\beta}) \exp \{ \E \bigg [  -\frac{1}{2} \sum_n    
 \bs{\beta}_n^2 \Tr(\bs{\Omega}_{n} ) 
  - 2 \bs{\beta}_n \Tr ( \bs{\Omega}_{n} \diag(  \bs{z}_n^\top   - \bs{X}^T \bs{w}_n^T) )   \bigg ]  \} \nonumber \tag{simplificiation }\\
  &\propto \exp  \{  - \frac{1}{2} \sum_n    
 \bs{\beta}_n^2 \underbrace{(\E[\tau_\beta] + \Tr(\E[\bs{\Omega}_{n}] ) )}_{:=\bs{\Phi}_n}
  - 2 \bs{\beta}_n \Tr ( \E[\bs{\Omega}_{n}] \diag \bigg (  \bs{z}_n^\top   - \E[\bs{X}^T] \E[\bs{w}_n^T] \bigg ) )   \} \nonumber \tag{definition of $p(\bs{\beta} | \tau_{\beta})$; applying expectation}\\
  &\propto \exp  \{ -\frac{1}{2} \sum_n  \bs{\Phi}_n \bigg (  \bs{\beta}_n - \frac{  \Tr ( \E[\bs{\Omega}_{n}] \diag \bigg (  \bs{z}_n^\top   - \E[\bs{X}^T] \E[\bs{w}_n^T] \bigg ) )  }{\bs{\Phi}_n} \bigg )^2 \} \nonumber \tag{completing the square; removing constants }
\end{align}

We can recognize the above expression as a product of univariate Gaussian distributions with mean and covariance denoted $\mu_n$ and $\sigma^2_n$. 
\begin{align}
    q^*(\bs{\beta}) = \prod_n \mathcal{N}(\mu_n, \sigma_d^2) \nonumber \\
    \mu_n = \sigma_n^2 \Tr ( \E[\bs{\Omega}_{n}] \diag \bigg (  \bs{z}_n^\top   - \E[\bs{X}^T] \E[\bs{w}_n^T] \bigg ) ) \And \sigma_n^{2} = \frac{1}{\E[\tau_\beta] + \Tr(\E[\bs{\Omega}_n])}
\end{align}

\subsection{Loading weights}

Recall $\bs{f}_n$ = $\bs{w}_n$ $\bs{X}_d + \bs{\beta}_n \bs{1}^\top $

\begin{align}
    q^*(\bs{W}) &\propto p(\bs{W}) \exp \{ \E_{q(-\bs{W})} [\log p(\bs{Y} | \bs{\Theta})]  \} \nonumber \\
    &\propto p(\bs{W}) \exp \{ \E \bigg [ -\frac{1}{2}\sum_n (\bs{z}_n - \bs{f}_n) \bs{\Omega}_{n}  (\bs{z}_n - \bs{f}_n)^\top \bigg ]  \} \nonumber \\
    &\propto p(\bs{W}) \exp \{ \E \bigg [  -\frac{1}{2}\sum_n -2  \bs{f}_n \bs{\Omega}_{n} \bs{z}_n^\top +   \bs{f}_n \bs{\Omega}_{n} \bs{f}_n^\top \bigg ]  \} \nonumber \tag{distribute and remove constants w.r.t $\bs{W}$}\\
    &\propto \exp \{ -\frac{1}{2} \bs{w_n} (\diag(\bs{\tau})) \bs{w_n}^\top \} \exp \{ \E \bigg [  -\frac{1}{2} \sum_n    
 \bs{w}_n \bs{X}  \bs{\Omega}_{n} \bs{X}^\top 
 \bs{w}_n^\top - 2  \bs{w}_n \bs{X}  \bs{\Omega}_{n}  \bs{z}_n^\top   \bigg ]  \} \nonumber \tag{removing constants w.r.t $\bs{W}$; $\hat{\bs{z}}_n^\top  = \bs{z}_n^\top - \bs{\beta}_n \bs{1} $ }\\
     &\propto \exp \{ \E \bigg [  -\frac{1}{2} \sum_n    
 \bs{w}_n ( \bs{X}  \bs{\Omega}_{n} \bs{X}^\top +  \diag(\bs{\tau})) 
 \bs{w}_n^\top - 2  \bs{w}_n \bs{X}  \bs{\Omega}_{n} \hat{\bs{z}}_n^\top   \bigg ]  \} \nonumber \tag{removing constants w.r.t $\bs{W}$}\\
    &\propto \exp \{ \E \bigg [  -\frac{1}{2} \sum_n   
 \bs{w}_n \bs{\Phi}_n \bs{w}_n^\top - 2  \bs{w}_n \bs{X}  \bs{\Omega}_{n} \hat{\bs{z}}_n^\top   + {\hat{\bs{z}}_{n}} \bs{\Omega}^{-1} \bs{X}^\top \bs{X} \bs{\Omega}_{n} \hat{\bs{z}}_n^\top -  {\hat{\bs{z}}_{n}} \bs{\Omega}^{-1} \bs{X}^\top \bs{X} \bs{\Omega}_{n} \hat{\bs{z}}_n^\top \bigg ]  \} \nonumber \tag{$\bs{\Phi}_n = ( \bs{X}  \bs{\Omega}_{n} \bs{X}^\top +  \diag(\bs{\tau}))$;  completing the square}\\
    &\propto \exp \{ \bigg [  -\frac{1}{2} \sum_n   
 \bigg ( \bs{w}_n - \bigg ( (E [\bs{\Phi}_n])^{-1} \E[\bs{X}] \E[\bs{\Omega_n} \hat{\bs{z}}_n^\top] \bigg )^\top   \bigg ) \E[\bs{\Phi}_n] \nonumber \\
 & \qquad \qquad \bigg ( \bs{w}_n - \bigg ( (E [\bs{\Phi}_n])^{-1} \E[\bs{X}] \E[\bs{\Omega_n} \hat{\bs{z}}_n^\top ] \bigg )^\top  \bigg )^\top \bigg ]  \} \nonumber \tag{removing constant;  applying expectations; completing the square; }
\end{align}

We can recognize the above expression as product of multivariate Gaussian distribution of random vectors $\bs{w}_n$ with mean and variance $\bs{m}_{n}$ and variance $\bs{V}_n$ 

\begin{align}
    q^*(\bs{W}) = \prod_n \mathcal{N}(\bs{m}_n, \bs{V}_n) \nonumber \\
    \bs{m}_n = \bs{V}_n (\E[\bs{X}] \E[\bs{\Omega_n} \hat{\bs{z}}_n^\top]) \And \bs{V}_n = (\E[\bs{\Phi}_n])^{-1}
\end{align}

\subsection{Precision Variables}

For the ARD precision variables,

\begin{align}
    q^*(\btau) &\propto  \exp \{ \E_{q(-\btau)} [\log p(\bY, \bTheta)]\}  \nonumber \\
    &\propto p(\btau) \exp \{ \E \left [  \log p(\bW | \btau) \right] \} \tag{no $\btau$ dependence}\\
    &= p(\btau) \exp \{ \E \left [ \sum_n \frac{1}{2}  \log |\diag(\btau) | - \frac{1}{2} \bw_n \diag(\btau) \bw_n^\top  \right] \} \tag{definition of $p(\bW| \btau)$} \\
    &= p(\btau) \exp \{ \E \left [ \frac{N}{2}  \sum_d \log \tau_d - \frac{1}{2} \sum_n  \bw_n \diag(\btau) \bw_n^\top  \right] \} \tag{definition of $p(\bW| \btau)$} \\
    &= p(\btau) \exp \{ \frac{N}{2}  \sum_d \log \tau_d   -\frac{1}{2} \sum_d \sum_n  \E \left [ \bw_{n,d}^2  \right] \tau_d + \const\}  \tag{taking expectations, rearranging} \nonumber  \\
    &\propto \exp \{ \sum_d (a_d - 1) \log \tau_d  - b_d \tau_d \} \exp \{ \frac{N}{2}  \sum_d \log \tau_d   -\frac{1}{2} \sum_d \sum_n  \E \left [ \bw_{n,d}^2  \right] \tau_d \}  \nonumber  \tag{definition of $p(\tau)$}\\
    &= \exp \{ \sum_d (a_d + \frac{N}{2} - 1) \log \tau_d  - (b_d + \frac{1}{2} \sum_d \sum_n  \E \left [ \bw_{n,d}^2  \right] )\tau_d \}.  \tag{rearranging} 
\end{align}

Recognizing the above as a  product of gamma distributions, we get 
\begin{align}
    q^*(\btau) = \prod_d \mathcal{G}(\hat{a}_d, \hat{b}_d) \qquad \text{where} \nonumber\\   
    \hat{a}_d = a_d + \frac{N}{2} \qquad \hat{b}_d = b_d + \frac{1}{2} \sum_n  \E \left [ \bw_{n, d}^2 \right]. \label{eq:appendix:cf-updates:tau} 
\end{align}

The update rule for the precision variable, $\tau_{\beta}$, is given as 
\begin{align}
    q^*(\tau_{\beta}) &\propto  \exp \{ \E_{q(-\tau_{\beta})} [\log p(\bY, \bTheta)]\}  \nonumber \\
    &\propto p(\tau_{\beta}) \exp \{ \E \left [  \sum_n \log p(\beta_n  | \tau_{\beta} ) \right] \} \tag{no $\tau_{\beta}$ dependence}\\
    &\propto p(\tau_{\beta}) \exp \{ \E \left [ \sum_n \frac{1}{2}  \log \tau_{\beta}  - \frac{1}{2} \beta_n^2 \tau_{\beta} \right] \} \tag{definition of $p(\beta_n| \tau_\beta)$} \\
    &= p(\tau_{\beta}) \exp \{ \frac{N}{2}  \log \tau_\beta   - \frac{1}{2} \sum_n  \E \left [ \beta_n^2  \right] \tau_\beta \}  \tag{taking expectations, rearranging} \nonumber  \\
    &= \exp \{  (c + \frac{N}{2} - 1) \log \tau_\beta  -( d +  \frac{1}{2} \sum_n  \E \left [ \beta_n^2  \right])  \tau_\beta   \} . \nonumber  \tag{definition of $p(\tau_\beta)$}
\end{align}

Recognizing the above as a product of gamma distributions, we get 
\begin{align}
    q^*(\btau_\beta) = \mathcal{G}(\hat{c}, \hat{d})  \qquad \text{where} \nonumber \\
    \hat{c} = c + \frac{N}{2} \qquad \hat{d}= d + \frac{1}{2} \sum_n  \E \left [ \beta_n^2  \right].
    \label{eq:appendix:cf-updates:tau-beta}
\end{align}

\subsection{Computing variational moments }
All of the variables except the dispersion parameter $r_n$ have a known closed form for their first and second moments. 
For $r_n$, we compute the its moments using an efficient gamma based sampler presented in \cite{he2019data} (see its Appendix C for details). 
Also for completeness, we review moments and other important facts of the less common P\'{o}lya-Gamma, P\'{o}lya-Inverse Gamma, and Power Truncated Normal distributions in \cref{sec:background:important-densities}

%% file: sections/appendix/elbo.tex
\vspace{.5cm}
\section{ELBO expression}\label{appendix:elbo}

\label{appendix:elbo-derivation}
The evidence lower bound (ELBO), denoted as $\mathcal{L}$ defined in \cref{eq:lowerbound} is given as follows 
\begin{align}
    \mathcal{L} &=  \E_{q(\bTheta)} \left [ \log p(\bY| \bTheta )\right] - \KL [q(\bTheta) \| p(\bTheta)]  \nonumber
\end{align}
where $\bTheta = \{ \bX, \bW, \bbeta, \btau, \tau_{\beta}, \{ \omega_{n,t}, \tau_{n,t}, \xi_{n,t}, \} \}$ is the set of all latent variables and $\KL$ represents Kullback-Leibler divergence measure. 

This  follows from Jensen's inequality,
\begin{align}
    \log p(\bY) %
    &= \log \E_{p(\bTheta)} \left[ p(\bY|\Theta)   \right] \nonumber\\
    &= \log \E_{q(\bTheta)}\left[  \frac{p(\bY|\bTheta)p(\bTheta)}{q(\bTheta)} \right] \nonumber\\
    &\geq \E_{q(\bTheta)} \left[ \log \frac{p(\bY|\bTheta)p(\bTheta)}{q(\bTheta)} \right] =: \mathcal{L}. 
\end{align}

In the variational EM algorithm presented in \cref{alg:ccGPFA}, we update  the variational distributions utilizing closed form solutions. However, to update the GP parameters, we directly optimize them (via automatic differentiation) using the ELBO formula.  
This requires further simplifications of the ELBO.
In the following, we show details of the simplifications.    
\begin{align}
    \mathcal{L} = \E_{q(\Theta)} [ p(\bs{Y}| \bs{\Theta}) ] - \KL [q(\bTheta) \| p(\bTheta) ]
\end{align}
We note that in the expression, the variational expectation term has no dependence of GP parameters. 
And by independence assumptions presented in \cref{eq:inference:variational-family},

the above KL term  decomposes as 
\begin{align}
    &\hspace{-1cm}\KL [q(\bTheta) \| p(\bTheta)]  \nonumber \\
    &= \sum_d  \KL[q(\bX_d \| p(\bX_d))] + \sum_{n,t} \KL[q(\omega_{n,t} \| p(\omega_{n,t}))] +  \sum_{n,t} \KL[q(\tau_{n, t}) \| p(\tau_{n,t} )] \nonumber \\
    &\qquad + \sum_{n,t} \KL[q(\xi_{n, t}) \| p(\xi_{n,t} )] + \sum_{n} \KL[q(\bw_n) \| p(\bw_n | \btau)] \nonumber \\
    &\qquad +\sum_n \KL[q(\beta_n) || p(\beta_n)] +  \sum_{d} \KL[q(\btau_d) \| p(\btau_d) ] + \KL[q(\tau_{\beta}) \| p(\tau_{\beta})]. \nonumber
\end{align}

From the equation above, we note that only the first $\KL$ term is dependent on GP parameters. Thus, further simplifying the KL using  divergence formula of two multivariate Gaussians distributions,   
\begin{align}
    \KL [q(\bX_d) \| p(\bX_d)] &= \KL[ q(\bX_d| \hat{\boldsymbol{\mu}}_d, \hat{\bK}_d) \| p(\bX_d | \boldsymbol{\mu}_d, \bK_d) ] \nonumber\\
    &= \frac{1}{2} \left[ \log |\bK_d| - \log|\hat{\bK}_d| - T + \Tr \{ \bK_d  ^{-1} \hat{\bK}_d \} + (\boldsymbol{\mu}_d - \hat{\boldsymbol{\mu}}_d )^\top \bK_d^{-1} (\boldsymbol{\mu}_d - \hat{\boldsymbol{\mu}}_d)  \right ]. \nonumber
\end{align}

We can now define an objective function to optimize our %
GP parameters $\{ \theta_d \}_{d=1}^D$.
\begin{align}
    \mathcal{L}(\theta_d) &= - \KL [q(\bX_d) \| p(\bX_d)] + \const  \tag{removing constants w.r.t $\theta_d$} \nonumber \\
    &= \E_{q(\bX_d)}[\log q(\bX_d)] + \E_{q(\bX_d)}[\log p(\bX_d)] + \const \tag{KL def.} \nonumber \\
    &= \E_{q(\bX_d)}[\log p(\bX_d)] + \const \tag{removing constants w.r.t $\theta_d$} \\
    &= -\frac{1}{2} \left (  \log |\bK_d| +  \E_{q(\bX_d)}[\bX_d] \bK_d^{-1} \E_{q(\bX_d)}[\bX_d^\top] +  \Tr(\hat{\bK}_d \bK_d) \right) + \const . \label{eq:appendix:elbo:dispersion-params} 
\end{align}
Recall $\bK_d$ denotes the prior covariance of the $d$-th latent process with kernel parameters $\theta_d$. In practice, we use PyTorch \cite{paszke2019pytorch} automatic differentiation engine to compute gradients with respect to the GP length scale parameters $\{\theta_d\}$. 
For a sparse GPFA variant \cref{subsection:scalable-inference}, $\bK_d$ would be an $M \times M$ prior covariance matrix instead of a large $T \times T$. 

%% file: sections/appendix/natural-gradients.tex
\section{Natural Gradients}
The sparse GPFA variant shown in \cref{subsection:scalable-inference} avoids the expensive $T \times T$ matrix. However, still in the variational updates it uses all the data i.e. spike counts of all neurons across time. Here we show how we can extend the accelerate the inference by using only a mini-batch of the data. Using mini-batch of the data we obtain a noisy natural gradient \cite{hoffman2013stochastic}.

Consider a random minibatch of time points $\mathcal{T}$. The natural gradients based on these time points are give below. These gradients are scaled to match the dataset size using the ratio of the total number of time steps to the minibatch size $\frac{T}{|\mathcal{T}|}$.

\label{appendix:natural-gradients}
\subsection{Weights}
The natural parameters for the loading weights of the $n$-th neuron are   
\begin{align}
    \boldsymbol{\eta}_1^{(n)} = \frac{T}{|\mathcal{T}|}\E[\bX] \E[\bOm_n] (\bz_n^\top  - \E[\beta_n] \boldsymbol{1}]) & \quad \boldsymbol{\eta}_2^{(n)} = -\frac{1}{2} ( \diag(\E[\btau]) + \frac{T}{|\mathcal{T}|} \E[\bX \E[\bOm_n] \bX^\top])
    \label{eq:appendix:natural-gradients:weights}
\end{align}

\subsection{Latent Processes}
The natural parameters for the $d$-th latent process are

\begin{align} 
    &\boldsymbol{\eta}_1^{(d)} =\frac{T}{|\mathcal{T}|} \bigg (  \sum_n \E[\bOm_n]  \E[\bw_{n,d}] \big( \E[\bz_n^\top]  - \E[\beta_n] \boldsymbol{1} - \sum_{d'\neq d}  \E[\bw_{n, d'}] \E[\bX_{d'}^\top]    \big )\bigg ) \nonumber \\
    &\boldsymbol{\eta}_2^{(d)} = -\frac{1}{2} \left ( \bK_d^{-1}  + \frac{T}{|\mathcal{T}|}\sum_n \E[w_{n, d}^2] \E[\bOm_n] \right ) \nonumber\\\label{eq:appendix:natural-gradients:Xd}
\end{align}

\subsection{Base Intensities}

The natural parameters for the base intensity of the $n$ neuron are given by 

\begin{align}
    \boldsymbol{\eta}_1^{(n)} = \frac{T}{|\mathcal{T}|}(\E[\bz_n] - \E[\bw_{n}] \E[\bX]  ) \E[\bOm_n]  \boldsymbol{1} & \quad
    \boldsymbol{\eta}_2^{(n)} = - \frac{1}{2} (\E[\tau_{\beta}] + \frac{T}{|\mathcal{T}|}\Tr(\E[\bOm_n]) \nonumber \label{eq:appendix:natural-gradients:beta} \\
\end{align}

%% file: sections/appendix/important-densities.tex
\section{Important Densities}
\label{sec:background:important-densities}
\citet{glynn2019bayesian} and \citet{he2019data} formally defined new distributions P\'{o}lya-Inverse Gamma(P-IG) and Power Truncated Normal distributions respectively. For completeness we include their important details in this section. We also review P\'{o}lya-Gamma distribution. 

\subsection{P\'{o}lya-Inverse Gamma (P-IG) }
\label{sec:background:P-IG}
The P\'{o}lya-Inverse Gamma (P-IG) distribution is defined with an infinite dimensional parameter vector $\bs{d} = \{ d_1, d_2, ...\}$ and a scalar ``tilting'' parameter $c$. %
The distribution of a P-IG random variable  $x$ with parameter $\bs{d}$ and $c=0$ is equivalent to an infinite convolution of the well known generalized inverse Gaussian (GIG) distribution, 
\begin{align}
   \PIG( x| \bs{d}, c=0) \stackrel{D}{=} \sum_{k=1}^\infty GIG \bigg (-\frac{3}{2}, \frac{1}{\sqrt{2}d_k}, 0 \bigg )  
\end{align}

The general class of $\PIG$($\bs{d}, 0$) is defined as exponential tilting of P-IG($\bs{d}, c$), similar to P\'{o}lya Gamma distribution \cite{polson2013bayesian},
\begin{align}
    \PIG(x| \bs{d}, c)   \propto \exp(-\frac{c}{2} x) \PIG(x| \bs{d}, 0) ,
\end{align}

and equivalently,
\begin{align}
    \PIG(x | \bs{d}, c) \stackrel{D}{=} \sum_{k=1}^\infty GIG \bigg (-\frac{3}{2}, 2 c^2, \frac{1}{2k^2}\bigg ) . 
\end{align}

Theorem 2 in \cite{he2019data} shows the first moment of the distributions available in closed form,
\begin{align}
    \E[x] = \frac{1}{2c} (\psi(c + 1) - \psi(1)).
\end{align}

\subsection{Power truncated normal (PTN)}
\label{sec:background:PTN}
The Power Truncated Normal distribution, PTN, is defined with three parameters $p, a > 0$ and $b \neq 0 $. For a PTN variable $x$, its unnormalized density is given as  
\begin{align}
    \PTN( x) \propto x^{p-1} e^{-ax^2 + bx}, x > 0.
\end{align}

\citet{he2019data} showed (in Appendix C) an efficient sampling algorithm from the distribution. 
This is important in the variational updates to compute the mean. 

\subsection{P\'{o}lya-gamma distribution }
The P\'{o}lya-gamma distribution is part of the class of infinite convolution of gamma distributions. For a P\'{o}lya-gamma  variable $\omega$ with distribution $\PG(b, 0)$ where $b$ denotes the shape parameter, the variable is equal in distribution with infinite sum of gamma variables.
\begin{align}
    \omega \stackrel{D}{=} \frac{1}{2\pi^2} \sum_{k=1}^\infty \frac{g_k}{(k - 1/2)^2} 
\end{align}
where $g_k \sim \Gamma(b, 1)$. \citet{polson2013bayesian} showed the distribution of the general class of $\PG(b, c)$ distribution can be expressed as exponential tilting of of $\PG$ ,
\begin{align}
    \PG(\omega| b, c) \propto \exp(-\frac{c^2}{2} x) p(\omega| b, 0).
\end{align}

For our purpose, in the variational updates it is important to compute the first moment of the distribution.  As shown in \cite{polson2013bayesian}, the first moment is given as 
\begin{align}
    \E(\omega) = \frac{b}{2c} \tanh(c/2).
\end{align}